\documentclass[journal]{IEEEtran}
\usepackage[colorlinks]{hyperref}
\ifCLASSINFOpdf
\else
\fi
\usepackage{amsmath}
\usepackage{algorithm}
\usepackage{amsmath} 
\usepackage{amssymb}
\usepackage{subcaption}
\usepackage{courier} 
\usepackage{graphicx} 
\usepackage{natbib} 
\usepackage{xcolor}
\usepackage{arydshln}
\usepackage{colortbl}
\usepackage{csquotes}
\usepackage{tabularx}
\usepackage{algpseudocode}
\definecolor{class1}{HTML}{00FFFF}
\definecolor{class2}{HTML}{069AF3}
\definecolor{class3}{HTML}{7FFF00}
\definecolor{class4}{HTML}{FF7F50}
\definecolor{class5}{HTML}{FF00FF}
\definecolor{class6}{HTML}{FFD700}
\definecolor{class7}{HTML}{008000}
\definecolor{class8}{HTML}{808080}
\definecolor{class9}{HTML}{FF4500}
\definecolor{class10}{HTML}{000080}
\definecolor{class11}{HTML}{800000}
\definecolor{class12}{HTML}{AAA662}
\definecolor{class13}{HTML}{FFA500}
\definecolor{class14}{HTML}{9A0EEA}
\definecolor{class15}{HTML}{DDA0DD}

\begin{document}
\title{Spatio-Temporal Joint Density Driven  Learning for Skeleton-Based Action Recognition}
% \title{\textcolor{blue}{STJD-CL: Spatio-Temporal Joint Density Driven Contrastive Learning for Skeleton-Based Action Recognition}}

\author{Shanaka~Ramesh~Gunasekara,~\IEEEmembership{Student,~IEEE}, Wanqing Li,~\IEEEmembership{Senior Member,~IEEE} , Philip Ogunbona,~\IEEEmembership{Senior Member,~IEEE}, Jack Yang,~\IEEEmembership{Senior Member,~IEEE.}% <-this % stops a space
\thanks{Shanaka Ramesh Gunasekara, Wanqing Li, Philip Ogunbona and Jack Yang are with the Advanced Multimedia Research Lab, University of Wollongong, Australia.}% <-this % stops a space
% \thanks{\textit{Corresponding Author: Wanqing Li. }}% <-this % stops a space
}


\markboth{IEEE Transactions on Biometrics, Behavior, and Identity Science}%
{Shell \MakeLowercase{\textit{et al.}}: Bare Demo of IEEEtran.cls for IEEE Journals}



\maketitle

\begin{abstract}
Traditional approaches in unsupervised or self-supervised learning for skeleton-based action classification have concentrated predominantly on the dynamic aspects of skeletal sequences. Yet, the intricate interaction between the moving and static elements of the skeleton presents a rarely tapped discriminative potential for action classification. This paper introduces a novel measurement, referred to as spatial-temporal joint density (STJD), to quantify such interaction. Tracking the evolution of this density throughout an action can effectively identify a subset of discriminative moving and/or static joints termed \textquote{prime joints} to steer self-supervised learning. A new contrastive learning strategy named STJD-CL is proposed to align the representation of a skeleton sequence with that of its prime joints while simultaneously contrasting the representations of prime and non-prime joints. In addition, a method called STJD-MP is developed by integrating it with a reconstruction-based framework for more effective learning.
%\textcolor{blue}{Moreover, the effectiveness of STJD is validated by incorporating it into a reconstruction-based method, referred to as STJD-MP. In this approach, the redundancy is reduced by masking prime joints, thus creating a more challenging self-supervisory task. This, in turn, encourages the encoder to learn more generalized representations.}
%that exploits the Spatio-Temporal Joint Density for skeleton-base action recognition. \textcolor{blue}{Furthermore,  prime and non-prime representations are contrasted to minimise the interference with the prime joints because of noise.} 
Experimental evaluations on the NTU RGB+D 60, NTU RGB+D 120, and PKUMMD datasets in various downstream tasks demonstrate that the proposed STJD-CL and STJD-MP improved performance, particularly by 3.5 and 3.6 percentage points over the state-of-the-art contrastive methods on the NTU RGB+D 120 dataset using X-sub and X-set evaluations, respectively. The code is available at \href{https://github.com/ShanakaRG/STJD-Spatio-Temporal-Joint-Density-Driven-Learning-for-Skeleton-Based-Action-Recognition.git}{STJD}.
\end{abstract}

\begin{IEEEkeywords}
Self-supervised learning, skeleton-based action recognition, spatio-temporal joint density.
\end{IEEEkeywords}

% \fi
\IEEEpeerreviewmaketitle

\section{Introduction}
\label{sec:intro}

%Human action recognition (HAR) is an active research area in computer vision with a wide range of real-world applications. 
\IEEEPARstart{W}{ith} the advances in pose estimation~\cite{Fang01} and depth sensing techniques, 3D skeleton-based action recognition has become an active research area.  Although supervised methods have shown impressive results ~\cite{Chen2021a,wang01,shanaka01,Guo01,shanaka02,Bruno}, they require a substantial amount of annotated data for effective learning. Self-supervised learning, which utilizes unlabeled data, has emerged as an alternative and promising approach~\cite{aimclr,chuankun02}.
%, which is often time-consuming and labour-intensive. Therefore, self-supervised 3D action representation learning, which utilizes a large amount of unlabeled data, has become a desired approach.

In general, self-supervised methods for action recognition can be categorized into two groups: reconstruction-based and contrastive learning-based. Reconstruction-based methods often employ an encoder-decoder architecture to learn representation via different pretext tasks, including the prediction of randomly masked frames or body parts~\cite{SkeletonMAE,mao01}, a jigsaw puzzle~\cite{ms2l}, and motion prediction~\cite{Cheng01}. Recently, masked autoencoders (MAE) have shown significant performance improvements in action recognition~\cite{SkeletonMAE,sjepa,Wenhan01,mao01}. Unlike traditional approaches that employ random masking~\cite{SkeletonMAE,sjepa},  the reconstruction of masked moving joints~\cite{mao01} yields promising performance. Due to the inherent structural dependencies between joints, missing static joints are relatively easy to recover compared to moving joints. In random masking, the inclusion of static joints—which can be easily recovered—renders the training objective less challenging. In contrast, moving joints are more difficult to recover, presenting a more challenging pre-training task that enforces the encoder to learn discriminative motion dynamics.

\begin{figure*}[tb]
  \centering
	\includegraphics[scale=0.7]{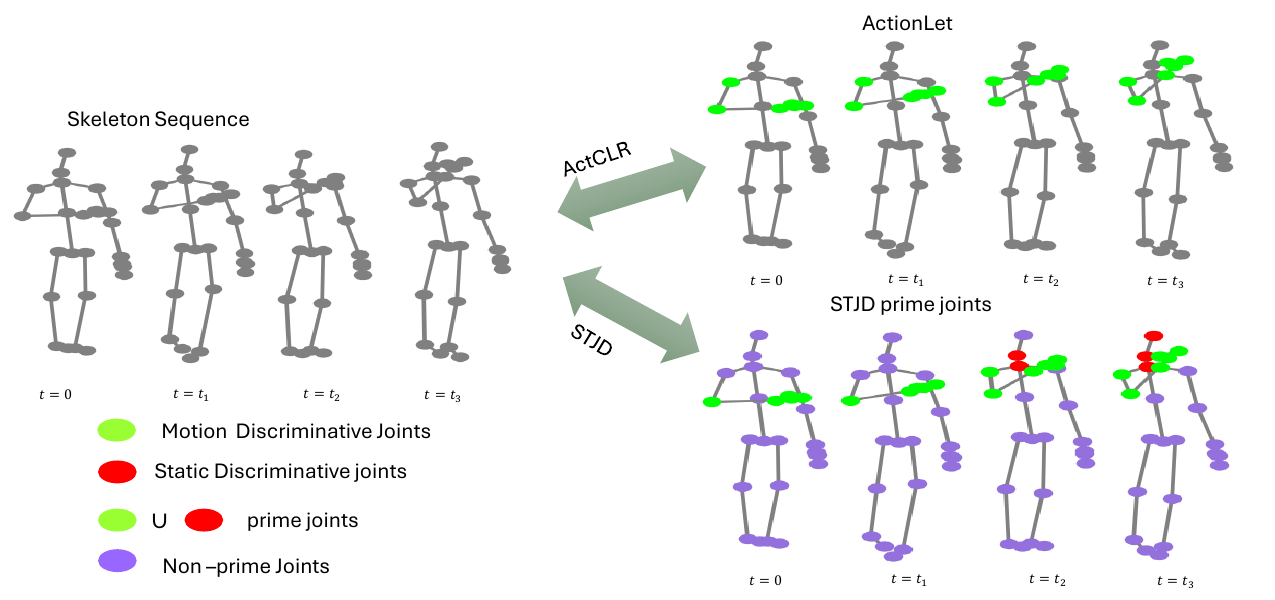}
	\caption{\textbf{Comparison of actionlets and prime joints for action Drink from NTU RGB+D 60}. An actionlet is selected on the basis of pre-defined parts using motion. In contrast, the prime joints are detected based on the evolution of the proposed STJD. The discriminative static head joints are failed to select in actionlet, but were successfully included in the prime joints. 
    % \textcolor{blue}{However, since no additional modules are introduced to the skeleton encoder, the inference computational complexity remains unchanged compared to the baseline model.}
    }	
  \label{fig:1}
\end{figure*}

%Prediction and/ or reconstruction of randomly masked frames or parts ~\cite{mao01,SkeletonMAE} are the most widely used pretext tasks so far.  
In contrastive learning, representations are learned by contrasting pairs of two sequences, transformed vs original or transformed vs transformed~\cite{aimclr,skelemixclr, ascal,Bulat}, aiming to align the pairs in a latent space~\cite{aimclr,skelemixclr}. The commonly used transformations include shear, spatial flip, axis mask, random crop, temporal flip, Gaussian noise, and Gaussian blur~\cite{aimclr,skelemixclr, ascal} and the choice of the transformation often affects the effectiveness of learning. Unlike the conventional approach to aligning the representations of two entire sequences, Lin et al.~\cite{actionlet} proposed to align the representation of a transformed sequence with the representation obtained only from moving parts, known as Actionlets, of a differently transformed sequence to improve the discriminative power of the learned representation. One of the issues with ActCLR~\cite{actionlet} is that its actionlets are restricted to {\it pre-defined and moving body-parts}.
%While the work of ActCLR~~\cite{actionlet} explores an interesting approach to guide contrastive learning using potentially discriminative parts, its actionlets are limited to moving body-parts that are predefined. 
Consequently, the actionlets are unable to capture the intricate interactions between moving and static joints, particularly when the moving and static joints are located in different predefined body parts, such as the head joints and hands in action \textquote{Drinking} as shown in Figure~\ref{fig:1}. Additionally, the predefined body parts may lead to the inclusion of irrelevant joints as part of the actionlets; for example, irrelevant spine joints in \textquote{Comb hair} are included along with head joints. These limitations have hindered the learning of discriminative representations. 

Previous methods commonly assume that discriminative information is carried by moving joints or predefined moving body parts, and they leverage these cues for contrastive learning or reconstruction tasks. However, these approaches do not explicitly model the interactions between moving and informative static joints. In contrast, the proposed STJD
% This paper introduces a novel measurement called spatial-temporal joint density (STJD) to 
quantify the interaction between moving and static joints which is essential for understanding joint dynamics. By tracking the spatial-temporal joint density throughout an action, a group of discriminative moving and static joints, referred to as \textquote{Prime Joints} \footnote{Prime joints are a group of action-related joints that include both moving and static joints. They are different from \textquote{active or dynamic joints} which often refer to moving joins.}, that are not confined to a predefined set of body parts, can be detected, as illustrated in Figure~\ref{fig:1}. It is important to note that the definition and approach to identifying prime joints using Spatial-Temporal Joint Density (STJD) differ from early works such as SMIJ~\cite{smij}, where the top $N$ moving joints are selected in an ad-hoc manner, and Actionlet~\cite{actionlet_old} which identifies informative joints in a supervised manner based on their ability to predict class labels. The method proposed in this paper, utilizing STJD, is an unsupervised learnable approach for detecting informative joints.

%\textcolor{blue}{Unlikely to SMIJ~\cite{smij}, which employs an ad-hoc method to select the top $N$ moving joints, and Actionlet~\cite{actionlet_old}, which identifies informative joints in a supervised manner based on their ability to predict class labels, STJD introduces a learnable, unsupervised approach for detecting informative joints.}

In addition, a contrastive learning framework, named STJD-CL, has been developed to align the representation of a skeleton sequence with that of its prime joints and to contrast the representations of the prime joints with those of the non-prime joints to further enhance learning. The STJD-MP has been developed through the integration of the STJD with a reconstruction-based framework to improve the learning process. 
%\textcolor{blue}{Further, prime and non-prime joint representations are contrasted to reduce interference from non-prime joints due to noise. It improves the discriminative power of prime joints and provides better steering for the learning process.} 
Experimental evaluations on the NTU RGB + D 60, NTU RGB + D 120 and PKUMMD datasets across various downstream tasks demonstrate the effectiveness of the proposed STJD, as well as the superiority of the STJD-CL and STJD-MP methods compared to existing state-of-the-art self-supervised approaches.

%Specifically, the kernel density estimation ~~\cite{parzen1,parzen3} is used to obtain STJD. The skeleton subregion named "Prime Region," which consists of prime joints, is acting as a driving force for self-supervised representation learning by maximizing the agreement between spatially averaged features of the whole skeleton and the prime region, as illustrated in Fig \ref{fig:1}. Using the STJD as a steering force, a novel contrastive learning framework is proposed as Spatio-Temporal Joint Density Driven Contrastive Learning referred to as STJD-CL  for skeleton based action recognition.  The proposed method is capable of learning semantically rich representations by discarding irrelevant information and retaining the most discriminative information for the downstream task. Extensive experiments in various downstream tasks with NTU RGB+D 60, NTU RGB+D 120, and PKUMMD datasets provide evidence of the effectiveness of the proposed method compared to the state-of-the-art.

The contributions of the proposed STJD can be summarized as follows. 
    \begin{itemize}
        \item A novel concept of spatial-temporal joint density (STJD) is introduced to quantify the dynamics of moving joints and their interaction with static joints simultaneously with \textit{learnable kernel density functions}.
        \item A method called STJD-CL is developed to detect a set of discriminative joints, referred to as prime joints, from the evolution of the STJD over an action. A prime-joint-driven contrastive learning framework (STJD-CL) is proposed for skeleton-based action recognition. In addition, STJD-MP is proposed by integrating the STJD with reconstruction-based methods. 
        \item Extensive experiments have been performed on three popular datasets in various downstream tasks to demonstrate the effectiveness of the proposed STJD and STJD-CL. 
    \end{itemize}
   
The remainder of this paper is organized as follows. In Section~\ref{related_work}, self-supervised representation learning frameworks and their adaptations for human activity recognition are reviewed. Section~\ref{method} describes the proposed STJD, STJD-CL, and STJD-MP models in detail. Section~\ref{expeiments} presents the benchmark datasets, experimental setups, main results and ablation studies. Finally, the paper is concluded with a discussion of the findings and potential future work in Section~\ref{discussion}.

\section{Related Works}
\label{related_work}
In this section, a brief review of self-supervised representation learning and its application to skeleton-based action recognition is presented. 

\subsection{ Self-supervised Representation Learning}
Early work on self-supervised learning is based on pre-tasks, such as reordering perturbed image patches~\cite{Doersch01}, predicting a permutation of multiple shuffled image patches called solving jigsaw puzzles~\cite{noroozi01}, Colourizing grayscale images~\cite{Zhang02}, and rotation prediction~\cite{Gidaris01,Zhai01} to learn image representation. In video representation learning,  prediction of the order of video frames~\cite{lee01,fernando02} is a common pretext task while solving space-time cubic puzzles inspired by jigsaw puzzles~\cite{noroozi01} is also used. In ~\cite{wang04}, learning of spatio-temporal feature representations is improved by regressing motion and appearance statistics across spatial and temporal dimensions. Subsequently, contrastive methods such as MoCo2~\cite{mocov2} and SimCLR~\cite{simclr} achieved notable improvement by effectively discerning positive pairs obtained by different transformations of the original sequences from negative ones in training a backbone network. To overcome the large memory requirement to store negative samples,  negative sample-free methods such as BYOL~\cite{byol}, SimSiam~\cite{simsiam} and Barlow Twins~\cite{barlowtwin} were introduced, and they use an asymmetric pair of networks to avoid feature collapse. These self-supervised strategies have been adopted in skeleton-based action recognition~\cite{ms2l,aimclr,pstl} as well. This paper adopts the MoCoV2~\cite{mocov2} architecture with STGCN~\cite{Yan01} as the encoder.

\subsection{Self-supervised Action Recognition}
Skeleton-based self-supervised action representation learning adopts both reconstruction and contrastive approaches. In the reconstruction-based approach, various pretext tasks have been employed to probe the action context inherent in skeleton sequences. Notably, LongTGAN~\cite{longTGaN} and Predict and Cluster (P\&C)~\cite{PnC}  employ a recurrent encoder-decoder architecture to reconstruct the input sequence and compare it with the original to learn a representation. Yang et al.~\cite{Siyuan01} colourized each joint based on its spatial and temporal order as a pretext to learn an action presentation.

Masked Autoencoders (MAEs)\cite{mae} have been used in learning 3D action representation~\cite{SkeletonMAE,macdiff}.  SkeletonMAE~\cite{SkeletonMAE} was a typical transformer-based MAE to reconstruct randomly masked joints. Instead of reconstructing the original sequences, subsequent approaches have explored alternative methods, such as denoising noisy sequences conditioned on the unmasked representation~\cite{macdiff, igm} and predicting masked joints in the embedding space~\cite{sjepa}. However, these methods often mask all joints indiscriminately. This issue was addressed in part by MAMP~\cite{mao01} by assigning higher masking probabilities to moving joints, but their interaction with static joints is overlooked.

The contrastive learning approach learns discriminative representations by enforcing similar representations between two transformed versions of a skeleton sequence while maintaining its dissimilarity to negative samples specified in a batch~\cite{thoker01} or mined from the samples stored in a memory bank~\cite{aimclr,cpm}. AS-CAL~\cite{ascal} and SkeletonCLR~\cite{skelemixclr} leverage momentum encoders to regularize the feature space together with different transformations of the original sequences. CrosSCLR~\cite{crossclr} improved SkeletonCLR~\cite{skelemixclr} by a cross-stream knowledge distillation strategy. To learn multiple different input skeleton representations, ISC~\cite{thoker01} introduced interskeleton contrastive learning. MS2L~\cite{ms2l} integrates reconstruction and contrastive strategies with multitask learning to improve the discriminative ability of the learned representation. Recently, AimCLR ~\cite{aimclr} introduced extreme augmentations, with the aim of learning general representations by providing harder contrastive pairs. ActCLR~\cite{actionlet} improves the learning of a discriminative representation by enforcing a similar representation between the transformed sequence and discriminative moving body parts, known as actionlets, of another differently transformed sequence. 

This paper takes a significant step further to remove all restrictions, such as moving joints and/or pre-defined parts, by introducing prime joints and a novel measurement called spatial-temporal joint density to detect the prime joints for guiding the learning.
%\textcolor{blue}{This paper generalizes the selection of fixed moving body parts to unconstrained discriminative moving and static joints, termed prime joints, and introduces a novel measurement called Spatiotemporal Joint Density to detect prime joints for guiding contrastive learning.}

% \textcolor{blue}{This paper expands from being only fixed moving body parts to unconstrained discriminative moving and static joints, termed prime joints, and introduces a novel measurement called Spatio-Temporal Joint Density to detect the prime joints for guiding contrastive learning.}

% This paper extends the actionlets from being fixed body-part and moving only to unconstrained discriminative moving and static joints (prime joints), and 

% Both AimCLR ~\cite{aimclr} and ActCLR ~\cite{actionlet} leveraged memory banks to store negative samples. 
% In PSTL ~\cite{pstl}, a negative sample-free triplet stream structure is proposed to exploit local relationships from partial skeleton sequences. 

\section{Proposed Method}
\label{method}

This section first introduces the concept of Spatial-Temporal Joint Density (STJD). The method for detecting prime joints using STJD is then described. Finally, the STJD-based self-supervised frameworks, STJD-CL and STJD-MP, are presented. 

\subsection{Spatial-Temporal Joint Density}
To effectively identify a group of discriminative joints, whether moving or static, a novel measurement called Spatial-Temporal Joint Density (STJD) is introduced. The proposal of STJD is motivated by the observation that both moving joints and their interactions with nearby static joints contribute to the discriminative information necessary for classifying an action. These moving and static joints form the set of  \textquote{Prime Joints}. STJD provides an effective means to quantify the interactions and influences among joints and, consequently, to detect the prime joints.
%\textcolor{blue}{ STJD aims not only to provide a means to identify the prime joints but also to quantify the interaction or influence among joints; the spatially closer, the higher the interaction. Note that prime joints may not necessarily limited to pre-defined body parts}

Let a skeleton sequence be represented as $X \in \mathcal{R}^{C \times V \times T }$  where $C$ is the number of channels, $V$ is the number of joints, and $T$ is the number of frames. 
%The objective is to select $V_{prime} \in V $ without being bound to body structure or motion dynamics.  
For a frame $X_t$ at time $t$, the interaction with a joint $r \in \{v_{t,1},v_{t,2} , ...,v_{t,V} \}$, where $v_{t,i}=\{v_{t,i,1}, v_{t,i,2}, \cdots, v_{t,i,C}\}$, by other joints is proposed to be measured by a density function $D_t(r)$ through the accumulation of influence from each joint. Assume that the influence of each joint on $r$ is represented as a kernel function $K(\cdot)$. In this paper, a Gaussian function is used as the kernel function $K(.)$, although other functions are also feasible. The distance between joints varies. For instance, the joints of a hand are located closely, while the joints of the knee and ankle are relatively distant. Therefore, hand joints tend to be consistently identified as having high STJD due to their close spatial proximity.  
To mitigate this, $D_t(r)$ is defined as a learnable function (ref. equation~\ref{eq3}) with \textit{sample-wise learnable bandwidth $h_i$} that is shared across channels. 
%\textcolor{blue}{Overcoming this, the $D_t(r)$ is defined as a learnable function (ref. equation~\ref{eq3}) by introducing a \textit{sample-wise learnable bandwidth $h_i$} that is shared across channels.}

% \textcolor{red}{remove this part 
%     \begin{align}
%         D_t(r) = \dfrac{1}{V}\sum_{i=1}^V  \prod_{j=1}^C \dfrac{1}{h_i}K  \left( \dfrac{r_{j}-v_{t,i,j}}{h_i}\right),
%     \label{eq2}
%     \end{align}
% where $r_{j}$ is the $j^{th}$ channel of joint $r$ and $h_i$ is a parameter to control the degree that joint $i$ interact with its surrounding joints.}

\begin{align}
        D_t(r) = \dfrac{1}{V}\sum_{i=1}^V  \prod_{j=1}^C \dfrac{1}{h_i}\dfrac{exp \big( -0.5 \big( \dfrac{r_{t,j}-v_{t,i,j}}{h_i}\big)^2\big)}{\sqrt{2 \pi}}  ,
    \label{eq3}
\end{align}
where $r_{t,j}$ is the $j^{th}$ channel of joint $r$ and $h_i$ is a learnable parameter that controls the degree to which the joint $i$ interacts with its surrounding joints.

The choice of STJD is motivated by its ability to capture both spatial and temporal joint interactions within a unified, learnable kernel function. Unlike purely spatial models that usually ignore the temporal evolution and pure spatial models that ignore the topology of joints, STJD integrates both aspects. This integration enables STJD to quantify not only the movement of individual joints but also their interactions, no matter whether they are static and moving. Moreover, the learnable kernel density function in STJD offers adaptability across various actions, improving robustness to noise and subtle variations in joint movements.

\subsection{Detection of Prime Joints via STJD }

$D_t(r)$ quantifies the interaction of the joint $r$ with all other joints at time $t$. To detect the prime joints, which include the moving joints as well as the static joints that interact with the moving joints, the temporal change $\Delta D(r)$ of the STJD is calculated as follows.
    \begin{align}
        \Delta D_{ t} (r) = D_{t+\delta t}(r) - D_t(r),
        \label{eq4}
    \end{align}
where $ D_t(r)$ and $D_{t+\delta t}(r)$ represent STJD  at time $t$ and $t+\delta t$, respectively. 
%Expanding Eq.~\ref{eq2} using Taylor series and then substituting it into Eq. \ref{eq4} it can be shown that $\Delta D_t(r)$ captures both moving and static joints. 
Without loss of generality,  a single channel dimension in Equation~\ref{eq3} with a Gaussian kernel function, $G(.)$, is considered.  %Expand Eq.~\ref{eq2} into a Talor series and keep the first three terms
Expanding $G(r;v_{t,i},h_i)$ around the joint $v_{t,k}$ into a Taylor series and retaining the first two terms,  $D_{t}(r)$ and $D_{t+ \delta t}(r)$ can be approximated as:    
\begin{equation}
    \begin{aligned}
        D_{t}(r) = {} & \ \dfrac{1}{V}\sum_{i=1}^V G(r;v_{t,i},h_i), \\
        \approx & \dfrac{1}{V}\sum_{i=1}^V \left[ G_{t,k} + \dfrac{\partial G}{\partial v} |_{v_{t,k}} (r-v_{t,k}) \right].
    \end{aligned}
    \end{equation}
    
    \begin{equation}
    \begin{aligned}
       D_{t+ \delta t}(r) = {} & \ \dfrac{1}{V}\sum_{i=1}^V G(r; v_{t+\delta t,i},h_i), \\ 
       \approx & \dfrac{1}{V}\sum_{i=1}^V \left[ G_{t+\delta t,k} + \dfrac{\partial G}{\partial v} |_{v_{t+\delta t,k}} (r-v_{t+\delta t,k})   \right], 
    \end{aligned}
    \end{equation}
where $G_{t,k}$ and $G_{t+\delta t,k}$ are the Gaussian kernel values at joint $v_{k}$ for frame $t$ and $t+\delta t$ respectively. The $\dfrac{\partial G}{\partial v} |_{v_{-,k}}$ is the derivative of the Gaussian function. Substituting the expansions into Equation~\ref{eq4} yields the change of STJD as

\begin{equation}
% \scriptsize
\begin{aligned}
        \Delta D_t(r) ={} & \ D_{t+\delta t}(r) - D_t(r),  \\
        =& \ \dfrac{1}{V}\sum_{i=1}^V G (r,v_{t+\delta t,i},h_i) - \dfrac{1}{V}\sum_{i=1}^V  G(r, v_{t,i},h_i) ,\\
        = & \ \dfrac{1}{V}\sum_{i=1}^V [ (G_{t+\delta t,k} - G_{t,k})  +  \\
        & \dfrac{\partial G}{\partial v} |_{v_{t,k}} \left( (r-v_{t+\delta t,k}) - (r-v_{t,k}) \right) ] , \\
        =&  \ \dfrac{1}{V}\sum_{i=1}^V [ (G_{t+\delta t,k} - G_{t,k}) + \\
        & (M(v_k)+M(r)) \dfrac{\partial G}{\partial v} |_{v_{t,k}}],   
\end{aligned}
\label{stjdchange}
\end{equation}

\begin{algorithm}
\caption{Spatial Temporal Joint Density}
\label{alg:stjd}
\textbf{Input:} 
\begin{itemize}
    \item A skeleton sequence in embedding space \(X\): \(X \in \mathbb{R}^{C \times V \times T}\).
    \item Learned bandwidth parameters \(\{h_i\}_{i=1}^V\), shared across channels.
\end{itemize}

\textbf{Output:} 
\begin{itemize}
    \item \(\Delta D_t(r)\), the spatiotemporal density change of joint \(r\) at time \(t\).
\end{itemize}

\begin{algorithmic}[1]
\State \textbf{Initialize:} For each joint \(r\) at time \(t=0\), set \(D_t(r) \gets 0\) and \(\Delta D_t(r) \gets 0\).
\For{\(t \gets 1\) to \(T\)}
    \For{\textbf{each} joint \(r \in \{v_{t,1}, \dots, v_{t,V}\}\)}
        \State \textbf{Compute} \(D_t(r)\) using:
        \[
          D_t(r) = \frac{1}{V} \sum_{i=1}^{V} 
          \prod_{j=1}^{C} \left(
            \frac{1}{h_i \sqrt{2\pi}} 
            \exp\Bigl(-0.5\Bigl(\frac{r_{t,j} - v_{t,i,j}}{h_i}\Bigr)^2\Bigr)
          \right)
        \]
    \EndFor
\EndFor
\For{\(t \gets 1\) to \( T-\delta t\)}
    \For{\textbf{each} joint \(r \in \{v_{t,1}, \dots, v_{t,V}\}\)}
        \State \textbf{Compute} the absolute temporal change:
        \[
            \Delta D_t(r) = \Bigl| D_{t+\delta t}(r) - D_t(r) \Bigr|
        \]
    \EndFor
\EndFor
\State \textbf{Normalize} the density change:
\[
\Delta D_t(r) = \mathrm{softmax}\bigl(\Delta D_t(r)\bigr)
\]
\State \textbf{Return} \(\{\Delta D_t(r)\}\) for all \(r\).
\end{algorithmic}
\end{algorithm}
where $M(v_k)=v_{t,k}-v_{t+\delta t,k}$ , $M(r)=r_{t}-r_{t+\delta t}$ , $G_{t,k}= \dfrac{1}{h_i}\dfrac{1}{\sqrt{2\pi}} exp \big( -0.5 \big( \dfrac{v_{t,k}-v_{t,i}}{h_i}\big)^2\big)$, and $G_{t+\delta t,k}= \dfrac{1}{h_i}\dfrac{1}{\sqrt{2\pi}} exp \big( -0.5 \big( \dfrac{v_{t+\delta t,k}-v_{t+\delta t,i}}{h_i}\big)^2\big) $. $M(.)$ represents the motion of joints, and $(G_{t+\delta t,k} - G_{t,k})$ quantifies the interaction of joint $v_k$ with other joints. Algorithm~\ref{alg:stjd} presents pseudo-code for the STJD calculation.
% In this derivation, assume that $\dfrac{\partial G}{\partial v} |_{v_{t,i}} = \dfrac{\partial G}{\partial v} |_{v_{t+\delta t,i}}$ since for a given joint $i$, the bandwidth $h_i$ is fixed and then the derivative of $G(v_i,\mu,h_i) $ will be equal for both case.  

%For the above simplification, it is assumed that  $\dfrac{\partial G}{\partial v} |_{v_{t,j}} = \dfrac{\partial G}{\partial v}_{v_{t+\delta t,j}}$ for small $\delta t$.
%for small $\delta t$ with constant bandwidth $h_i$ for a given joint $v_i$. 
According to the above, $\Delta D_t(r)$ between consecutive frames captures both the joint motion and the interactions among joints. Prime joints $J_{prime}$ comprise the joints whose $\Delta D_t (r)$ measurements are above a threshold $\beta$\footnote{$\beta$ is only applicable for the STJD-CL. Only $\Delta D_t(r)$ is used for the STJD-MP.} The remaining joints are classified as non-prime joints $J_{non-prime}$. The hyperparameter $\beta$ governs the model’s sensitivity and influences the detection of a discriminative set of joints. A lower $\beta$ results in a larger number of joints being detected as prime, capturing subtle motions/interactions but, if it is too small, increasing the risk of inclusion of noisy joints. Conversely, a too high $\beta$ threshold may overlook subtle joint movements

  \begin{figure*}[tb]
  \centering
	\includegraphics[scale=0.5,trim=0cm 3cm 0cm 3cm]{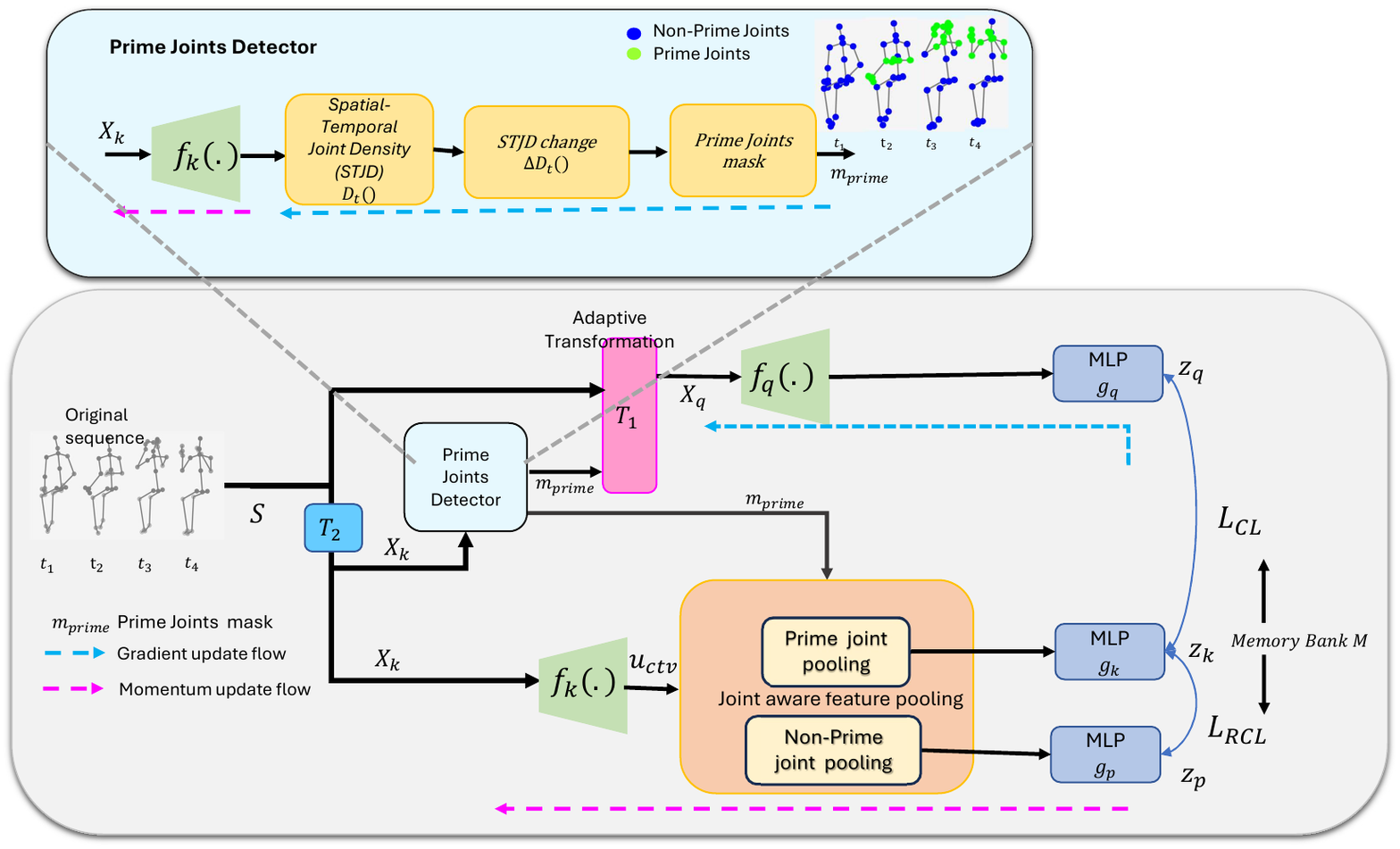}
	\caption{\textbf{The network architecture of the proposed STJD-CL}.Prime joints are detected via STJD. A two-stream network is used for contrastive learning, and the online stream is updated with gradients while the offline stream is updated via momentum. Adaptive transformation $\mathcal{T}_1$ is adopted from ~\cite{actionlet}. InfoNCE loss, $L_{CL}$, is defined to contrast the representation of the entire skeleton $X_q$ with that of the prime joints in $X_k$ and $L_{RCL}$ minimize the agreement between prime and non-prime representation. The STJD module is only used in the pertaining stage. Once pre-trained, the trained encoder $f_q(.)$ is used for the downstream tasks. Since no additional modules are introduced to the $f_q(.)$, the inference computational complexity remains unchanged compared to the baseline model.}	
  \label{fig:2}
\end{figure*}  

\subsection{STJD driven Contrastive Learning (STJD-CL)}
Figure~\ref{fig:2} shows the proposed STJD-CL contrastive learning framework, in which the learning is steered by the detected prime joints. Its architecture is similar to that of MoCoV2~\cite{mocov2}, consisting of two weight-shared encoders. The online query encoder $f_q(.)$ is updated using gradient descent, while the offline key encoder $f_k(.)$ is a momentum version of $f_q(.)$ with $\theta_k \leftarrow \alpha \theta_k + (1-\alpha)\theta_q$, where $\theta_k$ and $\theta_q$ are the parameters of the key and query encoders, respectively, and $\alpha$ is the momentum coefficient, which is typically close to 1. 

First, a transformed version of a skeleton sequence $S \in \mathcal{R}^{C_{in} \times T_{in} \times V}$, denoted as $X_k = \mathcal{T}_2(S)$, is mapped to an embedded space using the offline encoder $f_k(.)$. Then, the evolution of STJD  $\Delta D_t(r) \in \mathcal{R}^{1 \times T \times V}$ is calculated, where $\Delta D_{t=0}(r) =0$. By comparing $\Delta D_t(r)$ with a threshold value $\beta$, prime joints $J_{prime}$ are detected. The second transformed version is $X_q = \mathcal{T}_1(S) $, where $\mathcal{T}1(.)$ is an adaptive transformation adopted from ~\cite{actionlet}, and it is applied separately to $J_{prime}$ and $J_{non-prime}$ joints to preserve action semantics.

% The original action sequence $S$ fed into the offline encoder $f_k(.)$ and prime joint detection module is employed to generate prime region $X_{prime}$.

The online encoder maps $X_q$ into a latent representation $Z_q$ as $Z_q=g_q\big(GAP(f_q\big(\mathcal{T}_1(m_{prime}(\Delta D_t(h_i,S),S)\big))\big)$ , where $g_q(.)$ is an online projector and $GAP$ is global average pooling. In the offline encoder, latent representation $Z_k$ for the $J_{primt}$ of $X_k =  \mathcal{T}_2(S)$ is obtained by $Z_k = g_k \big(JAFP(f_k(\mathcal{T}_2(S),m_{prime}\big(\Delta D_t(h_i,S)\big)))\big)$, and the latent representation $Z_p$ for the $J_{non-prime}$ is obtained $Z_p = g_p \big(JAFP(f_k(\mathcal{T}_2(S),(1-m_{prime})\big(\Delta D_t(h_i,S)\big)))\big)$, where $g_k(.)$ and $g_p(.)$  are offline projectors, $\mathcal{T}_2$ is the transformation function and $JAFP$ is joint aware feature pooling which is performed over the $J_{prime}$(or $J_{non-prime}$) only to obtain the final feature representation $Z_k$(or $Z_p$).

\begin{equation}
\begin{aligned}
    JAFP(u_{ctv}) =  \sum_{t=1}^T\sum_{v=1}^V \dfrac{m_{-}}{\sum_{t=1}^T\sum_{v=1}^V m_{-}}  u_{ctv}
    \end{aligned}
\end{equation}
where $u_{ctv}$ is the output of the offline encoder and  $m_{-}$ is the mask for $J_{prime}$ or $J_{non-prime}$. 

$Z_k$s from different training samples are stored in a memory bank $M$ that is continuously updated using a first-in, first-out strategy. InfoNCE loss is used to contrast the latent representation of the entire skeleton $X_q$ with that of the prime joints in $X_k$ as follows. 

\begin{equation}
\scriptsize
    \begin{aligned}
        L_{CL} = \log \dfrac{exp(sim (Z_k, Z_q)/ \tau)}{exp(sim( Z_q, Z_k ) / \tau) + \sum _{j=1}^M exp(sim( Z_q, m_j) / \tau)}, 
        \end{aligned}
\end{equation}
%     \begin{equation}
% L_{CL} = \scalebox{0.81}{$\log \dfrac{\exp(\text{sim} (Z_k, Z_q)/ \tau)}{\exp(\text{sim}( Z_q, Z_k ) / \tau) + \sum _{j=1}^M \exp(\text{sim}( Z_q, m_j) / \tau)}$}
% \end{equation}
where $m_j$ are the negative samples mined from the stored samples in the memory bank $M$, and $\tau$ is the temperature hyperparameter. $sim(.)$ is the cosine similarity. In addition, to reduce the interference from the non-prime joints, a reversed contrastive loss $L_{RCL}$ between $Z_k$ and $Z_p$ is used to minimize their similarity. The final objective function to optimize the network is defined as
$L = L_{CL} + L_{RCL}.$

\subsection{STJD driven Mask Prediction(STJD-MP)}
To further validate the effectiveness of the proposed STJD, a reconstruction-based framework is developed. The MAMP~\cite{mao01} model was selected as a baseline, and its original motion-based masking strategy was replaced with a new strategy that masked out the prime-joints detected by the STJD. In the new strategy, higher masking probabilities were assigned to joints exhibiting larger spatial-temporal density variations, and then the model is trained to reconstruct the masked joints followed by the work~\cite{mao01}. This new training strategy is referred to as spatial-temporal joint density mask prediction (STJD-MP). 

%\textcolor{blue}{To further validate the effectiveness of the proposed STJD in a self-supervised setting, a reconstruction-based framework was developed. The MAMP~\cite{mao01} model was selected as a baseline, and its original motion-aware masking module—which masked out moving joints with higher probability—was replaced with the STJD module to maskout prime joints. Under STJD, higher masking probabilities were assigned to joints exhibiting higher spatial-temporal density variations. This modification yielded a new training strategy, referred to as spatial-temporal joint density mask prediction (STJD-MP). Following the same MAMP training, the masked joints were reconstructed using mean squared error (MSE). }
    
\section{Experimental Results}
\label{expeiments}
% This section sumurized the result obtained using the pre-trained encoder using the pretext task. 

\subsection{Datasets}

\textbf{NTU RGB+D 60 Dataset}~\cite{shahroudy01} comprises 56,578 action sequences categorized into 60 classes for skeleton-based action recognition. The dataset offers two evaluation protocols: Cross-Subject (X-sub) and Cross-View (X-view). In X-sub, subjects are split into training and test sets, while X-view uses samples from cameras 2 and 3 for training and camera 1 for testing. Each sequence provides 3D joint coordinates, with 25 body joints serving as nodes in the skeleton graph. 

% Initial features are derived from the 3D coordinates of these joints.
% \newline

\textbf{NTU RGB+D 120 Dataset}~\cite{Liu1905} extends the NTU RGB+D 60 dataset, offering a more extensive collection of 113,945 skeleton sequences across 120 action categories performed by 106 participants. It has two official evaluation protocols: Cross-Subject (X-sub) and Cross-Setup (X-set). In the X-sub protocol, the training set consists of actions from 53 subjects, while the remaining subjects contribute to the testing set. The X-set protocol categorizes data based on distinct camera setups, with training and validation sets collected from different setup IDs.
% \newline

\textbf{PKU-MMD Dataset}~\cite{Liu1703} captured using Kinect V2 sensors comprises almost 20,000 action sequences across 51 classes and involves 66 subjects. It consists of two subsets: Part I, an easier version for action recognition with 21,539 sequences, and Part II, a more challenging set of 6,904 sequences, characterized by substantial view variation and increased skeleton noise. The dataset facilitates cross-subject (X-sub) and cross-view (X-view) protocols for both subsets.

These datasets cover a wide range of actions including daily activities in homes and offices and interactions between humans, human and computers and humans and objects, They provide a comprehensive representation of real-world conditions by incorporating challenges such as self-occlusion and occlusion by objects. Furthermore, the use of varied camera setups—including differences in angles and distances— introduces diverse noise levels to the data, further enhancing their real-world applicability.

\subsection{Experimental Setup}

In our implementation of the STJD-CL, STGCN~\cite{Yan01} was used as the backbone encoder, following the same data preprocessing as that in AimCLR~\cite{aimclr} and ActCLR~\cite{actionlet} to ensure a fair comparison. The frame length of the action instances is set to 50 for all datasets, while the batch size for both pretraining and downstream tasks is 128, with the number of epochs set to 300. The hidden dimension of the STGCN backbone is reduced by a quarter to facilitate a fair comparison with the state-of-the-art (SOTA) results. The Adam optimizer is employed to train the network. The threshold value , $\beta ; 0< \beta < 1$,for the STJD is chosen empirically.
On NTU RGB+D 60 under x-view with joint modality, for $0.5 \leq \beta \leq 0.8 $, the recognition accuracy is $87.67\pm 0.0017$\% with a minimum of 87.44\% and a maximum of 87.93\% at $\beta = 0.65$. The same $\beta$ value is used for all the datasets, and SOTA performance is observed.
% \subsection{Linear evaluation }

For STJD-MP, transformer-based MAMP~\cite{mao01} backbone is used.
% \textcolor{blue}{Once the change of STJD is calculated using Equation~\ref{stjdchange}, a higher masking to the joints with higher density changes and model is trained to reconstruct the masked joints by following the work~\cite{mao01}}. 
The sequence length is set to 120 frames, and 400 epochs were used for pre-training. To ensure fairness in comparison, all other model components and hyperparameters remained unchanged.

\textbf{Linear evaluation:}

In the linear evaluation, a linear classifier $\phi(.)$ is attached to the encoder $f_q(.)$ to classify the extracted features. The encoder $f_q(.)$ remains fixed throughout the linear evaluation protocol, and  $\phi(.)$ is updated based on recognition accuracy. Table~\ref{tab01} shows the results and compares the improvement of STJD-CL over the baseline ActCLR~\cite{aimclr} for all three streams: Joint, Bone, and Motion. It can be seen that STJD-CL outperformed ActCLR in all cases. Table~\ref{tab02} summarizes the comparison of STJD-CL with other SOTA methods. The three-stream ensemble model is used to obtain the best results. STJD-CL outperformed ActCLR by $3.5$ percentage and $3.6$ percentage on the NTU RGB+D 120 dataset using X-sub and X-set evaluations, respectively. 

As shown in Table~\ref{tab01}, STJD-MP consistently outperformed the baseline MAMP across all evaluated datasets. On the NTU RGB+D 60 dataset, improvements of 1.1 percentage points and 0.5 percentage points were achieved over MAMP under the X-view and X-sub protocols, respectively. Moreover, as presented in Table~\ref{tab02}, STJD-MP achieved performance comparable to other leading reconstruction-based methods across multiple benchmarks. Notably, STJD-MP attained a new state-of-the-art result on the NTU RGB+D 120 dataset under the X-set evaluation, surpassing MacDiff~\cite{macdiff} by 0.2 percentage points.

\begin{table}[t]
    \caption{ Linear evaluation results of the proposed STJD-CL and STJD-MP in comparison with their baseline ActCLR and MAMP respectively on the NTU RGB+D 60, and NTU RGB+D 120 datasets.
  }
  
  \centering
  \begin{tabular}{l |l l|l l}
    % \toprule
    
     Models    & \multicolumn{2}{c}{NTU 60 (\%)} & \multicolumn{2}{|c}{ NTU 120 (\%)} \\
     \cline{2-5} 
    
                           &   X-sub &  X-view &  X-sub  &  X-set \\
    \cline{1-5} 
    
    % \midrule
    % AimCLR ~\cite{aimclr}        &  Joint (J)   & 74.3  & 79.7   & 63.4   & 63.4\\
    \textit{Stream: J } & & & & \\
    ActCLR~\cite{actionlet}       & 80.9  & 86.7   & 69.0   & 70.5\\
    STJD-CL (ours)             & 82.3   & 87.9  & 70.5  &  72.8 \\
    \cline{1-5} 
    % \midrule
    % AimCLR ~\cite{aimclr}        &  Bone (B)    & 73.2  & 77.0   & 62.9   & 63.4\\
    \textit{Stream: B}& & & & \\
    ActCLR~\cite{actionlet}              & 80.1  & 85.0   & 67.8   & 68.2\\
    STJD-CL (ours)               & 81.3  & 86.1   & 71.4  & 71.8\\
    \cline{1-5} 
    % \midrule
    % AimCLR ~\cite{aimclr}        &  Motion (M)  & 66.8  & 70.6     & 57.3   & 54.4\\
    \textit{Stream: M} & & & & \\
    ActCLR~\cite{actionlet}           & 78.6  & 84.4     & 68.3   & 67.8\\
    STJD-CL (ours)              & 81.4 & 85.8 & 71.4  & 71.7\\
    \cline{1-5} 
    % \midrule
    % 2s-CrosSCLR Joint + Motion 74.5 82.1
    % 3s-AimCLR ~\cite{aimclr}        &  J+B+M  & 78.9 & 83.8     & 68.2   & 68.8\\
    \textit{Stream: J+B+M} & & & & \\
    3s-ActCLR~\cite{actionlet}            & 84.3  & 88.8     & 74.3   & 75.7\\
    \textbf{3s-STJD-CL   (ours) }             & \textbf{85.9} & \textbf{90.0}  & \textbf{77.8}   &\textbf{79.3}\\
    \cline{1-5} 
    \textit{Stream: J } & & & & \\
    MAMP~\cite{mao01} & 84.9 & 89.1 & 78.6 & 79.1 \\
    \textbf{STJD-MP (ours)} & \textbf{85.4} &\textbf{90.2} & \textbf{79.1} & \textbf{80.4} 
  %   \midrule

  % \bottomrule
  \end{tabular}

  \label{tab01}
\end{table}

\begin{table*}[t]
  \caption{ Comparison of action recognition results with the linear evaluation on the NTU RGB+D and PKUMMD Datasets. * indicates the reproduced results using the released code. \textbf{Bold} and \underline{underlined} indicate the best and second
best results, respectively.}  
  
  \centering
  \begin{tabular}{l | l |c |c |c |c |c|c}
    % \toprule
      Models & Backbone & \multicolumn{2}{c|}{NTU 60}  & \multicolumn{2}{c|}{ NTU 120} &\multicolumn{2}{c}{ PKU MMD} \\
     \cline{1-8} 
                       &    &  X-sub  &  X-view  &  X-sub &  X-set  & Part I & Part II\\

    % \midrule
    \cline{1-8} 
    {\cellcolor{gray!20}}\textcolor{gray}{\textit{Reconstruction-based Single-stream: J}} & {\cellcolor{gray!20}}&{\cellcolor{gray!20}} &{\cellcolor{gray!20}} &{\cellcolor{gray!20}} & {\cellcolor{gray!20}}& {\cellcolor{gray!20}}& {\cellcolor{gray!20}} \\
    P \& C~\cite{PnC}      &    GRU    & 50.7 & 76.1  & - & -& - & - \\
   SkeletonMAE \cite{Wenhan01}& Transformer &74.8 & 77.7 & 72.5 & 73.5 & - & 36.1 \\
    MAMP~\cite{mao01}  &Transformer & 84.9 & 89.1 & 78.6 & 79.1& 92.2  \\
    S-JEPA~\cite{sjepa}  &Transformer &85.3 &89.8 &\textbf{79.6} &\textbf79.9 &92.2 \\
    % IDM & & & & &\\
    MacDiff~\cite{macdiff}  &Transformer & \textbf{86.4} & \textbf{91.0}& \underline{79.4} & \underline{80.2}& 92.8 \\
    \textbf{STJD-MP} (ours) &Transformer & \underline{85.4} & \underline{90.2} & 79.1 & \textbf{80.4}&   \\

    \cline{1-8} 
    % \textbf{\textit{Contrastive models}} & & & & & &  \\
    {\cellcolor{gray!20}}\textcolor{gray}{\textit{Contrastive Single-stream: J}} &{\cellcolor{gray!20}} & {\cellcolor{gray!20}}&{\cellcolor{gray!20}} &{\cellcolor{gray!20}} & {\cellcolor{gray!20}}&{\cellcolor{gray!20}}&{\cellcolor{gray!20}}  \\
    % LongT GAN ~\cite{longTGaN}      & 39.1 & 48.1 & - & - & 67.7 & 26.0\\
    MS2L ~\cite{ms2l}           &   GRU  & 52.6& -& - & -& 64.9 & 27.6\\
    AS-CAL ~\cite{ascal}         & LSTM  & 58.5 & 64.8 & 48.6& 49.2 & - & -\\
    PCRP ~\cite{Shihao01}      &   GRU  & 53.9 & 63.5  & 41.7&  45.1& - & -\\
   
    SeBiReNet ~\cite{nie01}     &  GRU  & -    & 79.7 & - & -& - & -\\
    SkeletonCLR ~\cite{crossclr}  & GCN & 68.3 & 76.4& 56.8 & 55.9 & 80.9 & 36.0\\
    ISC ~\cite{thoker01}        &  GCN  & 76.3 & 78.6 & 67.9  & 67.1 & 80.9 & 36.8\\
    AimCLR ~\cite{aimclr}      & GCN     & 74.3  & 79.7   & 63.4  & 63.4& 84.7 & 38.2\\
    GL-Transformer ~\cite{glformer} &Transformer & 76.3 & 83.8 & 66.0 & 68.7& - & - \\
    PSTL ~\cite{pstl}            &  GCN & 77.3 & 81.8  &  69.2 & 67.7 & 88.4 & 49.3\\
    CPM ~\cite{cpm}              &   GCN& 78.7 & 84.9 & 68.7 & 69.6 & - & -\\
    SkeleMixCLR ~\cite{skelemixclr} & GCN& 80.7 & 85.5 & 69.0 & 68.2 & - & -\\ 
    ActCLR ~\cite{actionlet}     & GCN   & 80.9  & 86.7   & 69.0   & 70.5 & - & -\\
    \textbf{STJD-CL } (ours)      & GCN         & \textbf{82.3}  & \textbf{87.9}   & \textbf{70.5}   &\textbf{72.8}& \textbf{90.6} & \textbf{51.5}\\
    \cline{1-8} 
    % \midrule
   
    {\cellcolor{gray!20}}\textcolor{gray}{\textit{Contrastive Multi-stream: J+B+M} }  &{\cellcolor{gray!20}} &{\cellcolor{gray!20}} &{\cellcolor{gray!20}} & {\cellcolor{gray!20}}&{\cellcolor{gray!20}} & {\cellcolor{gray!20}}& {\cellcolor{gray!20}}\\
    % 3s-Colorization ~\cite{Siyuan01}    & 75.2 & 83.1 & - & - & - & -\\
    3s-SkeletonCLR ~\cite{crossclr}   & GCN   & 75.0 & 79.8     &  60.7      &  62.6  & - & -\\ 
    3s-CrosSCLR ~\cite{crossclr}      & GCN   & 77.8 & 83.4     & 67.9   & 66.7 & 84.9 & 21.2\\
    3s-AimCLR ~\cite{aimclr}        & GCN     & 78.9 & 83.8     & 68.2   & 68.8& 87.8 & 38.5\\
    3s-PSTL ~\cite{pstl}           & GCN    & 79.1 & 83.8  & 62.2 &70.3 & \underline{89.2} & \underline{52.3} \\ 
    3s-SkeleMixCLR  ~\cite{skelemixclr}  & GCN& 82.7 & 87.1 & 70.5 & 70.7 & - & -\\
    3s-CPM ~\cite{cpm}             &   GCN  & 83.2 & 87.0  & 73.0 & 74.0 & - & -\\
    3s-ActCLR ~\cite{actionlet}     & GCN     & \underline{84.3}  & \underline{88.8}     & \underline{74.3}   & \underline{75.7} & - & -\\
    \textbf{3s-STJD-CL} (ours)        & GCN          & \textbf{85.9}  & \textbf{90.0}   &\textbf{ 77.1}   &\textbf{79.3}  & \textbf{93.2}& \textbf{55.3}\\
    % \midrule

  % \bottomrule
  \end{tabular}

  \label{tab02}
\end{table*}

% \subsection{Supervised fine-tune evaluation}
\textbf{Supervised fine-tune evaluation:}
A linear classifier is added to the unsupervised pre-trained encoder $f_q(.)$, and both the encoder and the linear classifier are further fine-tuned on a labeled training dataset. The results are compared with the SOTA methods and are shown in Table \ref{tab03}. STJD-CL exhibits superior performance, surpassing all SOTA methods and even outperforming the fully supervised STGCN~\cite{Yan01} by more than $3$ percentage points in all evaluations on both the NTU RGB+60 and NTU RGB+D 120 datasets.

Under supervised fine-tuning, STJD-MP outperformed all existing SOTA methods, except in the NTU RGB+D 120 X-sub evaluation setting. The proposed STJD-MP increased accuracy from 97.6\% to 97.8\% on the NTU RGB+D 60 X-view dataset.

The outcomes demonstrate the effectiveness of the proposed STJD on STJD-CL and STJD-MP.

\begin{table}[htb]
 \caption{ Supervised fine-tune evaluation results on the NTU RGB+D 60, and NTU RGB+D 120 datasets. \textbf{Bold} and \underline{underlined} indicate the best and second
best results, respectively.
  }  
  
  \centering
  \begin{tabular}{@{}l |cc|cc}
    % \toprule
    
     Models     & \multicolumn{2}{c}{NTU 60 (\%)} & \multicolumn{2}{|c}{ NTU 120 (\%)} \\

                           &   X-sub &  X-view &  X-sub  &  X-set \\

    \cline{1-5} 
    % \midrule
    {\cellcolor{gray!20}}\textbf{\textcolor{gray}{Contrastive methods}} &{\cellcolor{gray!20}} & {\cellcolor{gray!20}}& {\cellcolor{gray!20}}& {\cellcolor{gray!20}}\\
    \textcolor{gray}{\textit{Single Stream: J}} & & & & \\
    STGCN        & 81.5 & 88.3   & 70.7    & 73.2 \\
    % SkeletonCLR ~\cite{crossclr} &  Joint (J)   & 82.2  & 88.9   & 73.6    & 75.3 \\
    AimCLR      & 83.0  & 89.2   & 77.2   & 76.1 \\
    % PSTL ~\cite{pstl}            &  Joint (J)   & 84.5  & 92.0   & 78.6  & 78.9 \\
    CPM                 & 84.8  & 91.1   & 78.4   & 78.9 \\
    ActCLR       & \underline{85.8}  & \underline{91.2}   & \underline{79.4}   & \underline{80.9} \\
    
    \textbf{STJD-CL} (ours)              & \textbf{86.6}  & \textbf{92.4}  &  \textbf{80.3}  &  \textbf{82.0} \\
    % \midrule
    \cline{1-5}  
    % \cdashline{1-5} 
    \textcolor{gray}{\textit{Multi stream: J+B+M}} & & & & \\
    % 2s-CrosSCLR Joint + Motion 74.5 82.1
    3s-STGCN           & 85.2 & 91.4  & 77.2 & 77.1 \\
    3s-CrosSCLR       & 86.2 & 92.5  & 80.5 & 80.4 \\
    3s-AimCLR         & 86.9 & 92.8  & 80.1   & 80.9\\
    % 3s-PSTL ~\cite{pstl}            &  J+B+M  & 87.1 & 93.9 & 81.3 & 82.6 \\
    3s-ActCLR      & \underline{88.2}  & \underline{93.9} &\underline{ 81.6 }  & \underline{81.2}\\
    \textbf{3s-STJD-CL}   (ours)             & \textbf{89.3 } & \textbf{94.8}   & \textbf{83.5}   &\textbf{86.8}\\
    \cline{1-5} 
    {\cellcolor{gray!20}} \textbf{\textcolor{gray}{Reconstruction methods}} &{\cellcolor{gray!20} }&{\cellcolor{gray!20}} &{\cellcolor{gray!20}} &{\cellcolor{gray!20}} \\
    \textcolor{gray}{\textit{Single Stream: J}} & & & & \\
    SkeletonMAE & 88.5&94.7 &87.0 &88.9 \\
    MAMP & \underline{93.1}& 97.5&90.0 & \underline{91.3} \\
    Macdiff &92.7 &97.3 & -& -\\
    S-JEPA & \underline{93.1}&\underline{97.6} &\textbf{90.3} & \underline{91.3} \\
    \textbf{STJD-MP}(ours) &\textbf{93.5} & \textbf{97.8} & \underline{90.1} & \textbf{91.6}\\
    % \midrule
    % 2nd-level heading & {\bf 2.1 Printing Area} & 10 point, bold\\
    % 3rd-level heading & {\bf Headings.} Text follows \dots & 10 point, bold\\
    % 4th-level heading & {\it Remark.} Text follows \dots & 10 point, italic\\
  % \bottomrule
  \end{tabular}
 
  \label{tab03}
\end{table}

% \subsection{Semi-supervised evaluation}
\textbf{Semi-supervised evaluation:}
The effectiveness of the proposed STJD-CL is also evaluated in a semi-supervised setting. The encoder $f_q(.)$ is initially trained in an unsupervised manner, followed by fine-tuning with a classification layer appended as the last layer. Fine-tuning is conducted using 1\% and 10\% labeled data on the NTU RGB+D 60 and PKUMMD datasets for both X-Sub and X-View evaluation settings. The results in Table~\ref{tab04} demonstrate the effectiveness of the proposed method, surpassing the state-of-the-art by a large margin. Specifically, on the NTU RGB+D 60 dataset using X-Sub and X-View protocols, STJD-CL outperformed ActCLR~\cite{actionlet} by $2.7$ and $2.8$ percentage points, respectively, when 1\% of the labeled data was used in fine-tuning. STJD-CL outperformed PSTL~\cite{pstl} by 6.8\% on the PKUMMD Part-I dataset with 1\% of labeled data for fine-tuning. 

\begin{table}
   \caption{ Semi-supervised performance of STJD-CL and its comparison with the SOTA contrastive methods.
  } 
  \centering
  \begin{tabular}{l |cccc}
    % \toprule
    
     Models    & \multicolumn{2}{c}{NTU 60 (\%)} & \multicolumn{2}{c}{ PKUMMD(\%)} \\
     % \cline{3-4}
    
                           &   X-sub &  X-view & Part I  &  Part II \\

     \cline{1-5} 
     \textcolor{gray}{\textit{1\% labels}} & & & & \\
    % \midrule
    % LongT GAN          & 35.2 & -   & 35.8   & 12.4 \\
    % SkeletonCLR ~\cite{crossclr} &  Joint (J)   & 82.2  & 88.9   & 73.6    & 75.3 \\
    MS$^2$L           & 33.1  & -   & 36.4   & 13.0 \\
    % PSTL ~\cite{pstl}            &  Joint (J)   & 84.5  & 92.0   & 78.6  & 78.9 \\
    % ISC ~\cite{}              &  1   & 35.7  & 38.1   & 37.7   & - \\
    3s-CrosSCLR  & 51.1 & 50.0 & 49.7 & 10.2 \\
    3s-AimtCLR       & 54.8  & 54.3   & 57.5   & 15.1 \\
    % CPM                & 56.7 & 57.5 & - & - \\
    3s-ActCLR  & 64.8 & 65.6 & - & - \\
    3s-PSTL              & -  & -   & 62.5  & 16.9 \\
    \textbf{3s-STJD-CL} (ours)              & \textbf{67.5}  & \textbf{68.4}  &  \textbf{69.3}  &  \textbf{18.0} \\
    \cline{1-5} 
    \textcolor{gray}{\textit{10\% labels} }& & & & \\
    % \midrule
    % \cline{1-6} \\
    % 2s-CrosSCLR Joint + Motion 74.5 82.1
     % LongT GAN         & 62.0 & -   & 69.5   & 25.7 \\
    % SkeletonCLR ~\cite{crossclr} &  Joint (J)   & 82.2  & 88.9   & 73.6    & 75.3 \\
    MS$^2$L           & 65.2  & -   & 70.3  & 26.1 \\
    % PSTL ~\cite{pstl}            &  Joint (J)   & 84.5  & 92.0   & 78.6  & 78.9 \\
    % ISC ~\cite{}              &  1   & 35.7  & 38.1   & 37.7   & - \\
    % CPM                  & 73.0 & 77.1 & - & - \\
    3s-CrosSCLR   & 74.4 & 77.8 & 82.9 & 28.6 \\
    3s-AimtCLR       & 78.2  & 81.6   & 86.1   & 33.4 \\
    3s-ActCLR   & 81.7& 85.8& - & - \\
    3s-PSTL            & -  & -   & 86.9 & 42.0 \\

    \textbf{3s-STJD-CL} (ours)              & \textbf{82.5}  & \textbf{88.0}  &  \textbf{88.2}  &  \textbf{42.9} \\
    % \midrule

  % \bottomrule
  \end{tabular}

  \label{tab04}
\end{table}

To further validate the effectiveness of STJD, the model STJD-MP was evaluated under a semi-supervised setting. The classification layer and pre-trained encoder were fine-tuned together using only a small fraction of the training data (1\% and 10\%). The same supervised fine-tuning protocol was employed, and the average accuracy over five runs was reported in Table~\ref{tab04_2} to facilitate a fair comparison with SOTA reconstruction-based methods. Under these conditions, competitive performance was consistently demonstrated across datasets, and SOTA results were achieved on the NTU RGB+D X-sub benchmark using only 1\% and  10\% of the available training data.

\begin{table}
\caption{Semi-supervised performance of STJD-MP and its comparison with the SOTA reconstruction-based methods. \textbf{Bold} and \underline{underlined} indicate the best and second best results, respectively.}
    \centering
    \begin{tabular}{l |cc|cc}
    
     \cline{1-5} 
     Models    & \multicolumn{4}{c}{NTU RGB+D 60 (\%)}  \\
      & \multicolumn{2}{c|}{\cellcolor{gray!20} \textit{1\% labels}} & \multicolumn{2}{c}{\cellcolor{gray!20} \textit{10\% labels}} \\
    
                           &   X-sub &  X-view & X-sub & X-view \\

     \cline{1-5}

    MAMP  & 66.0 & 68.7 & 88.0 & 91.5 \\
    MacDiff & 65.6 & \textbf{77.3} & \underline{88.2} & \textbf{92.5} \\
    S-JEPA & \underline{67.5} & 69.1 & \textbf{88.4} & 91.4 \\
    \cline{1-5} 
    \textbf{STJD-MP} (ours)              & \textbf{69.5}  & \underline{70.2} & \textbf{88.4} & \underline{92.2} \\
    \cline{1-5} 

  \end{tabular}

  \label{tab04_2}
\end{table}

% \subsection{Transfer learning}
\textbf{Transfer learning:}
The pre-trained encoder $f_q(.)$ on the NTU RGB+D 60 dataset was employed for linear evaluation using the X-sub protocol on the PKUMMD dataset. The results are presented in Table~\ref{tab05}. The results exceed the SOTA by $1.4$ percentage and $5.8$ percentage on the PKUMMD I and PKUMMD II datasets, respectively. This demonstrates the generalizability of the proposed framework to various downstream tasks, resulting from the effective detection of the prime joints using STJD.

\begin{table}
    \caption{ Transfer learning performance of STJD-CL on the PKUMMD Dataset: linear evaluation with the encoder pre-trained on NTU RGB+D 60.
  }
  \centering
  \begin{tabular}{@{}l |cc}
    % \toprule
    
     Models    &   \multicolumn{2}{|c}{ PKUMMD(\%)} \\
                           & Part I  &  Part II \\
    % \midrule
    \cline{1-3} 
     % LongT GAN         & -   & 44.8 \\
    % MS$^2$L         & -  & 45.8 \\
    3s-CrosSCLR   & - & 51.3 \\
    3s-AimtCLR         & 85.6   & 51.6 \\
    3s-ActCLR & 90.0 & 55.9 \\
    % CPM ~\cite{cpm}                 & - & - \\
    % 3s-PSTL ~\cite{pstl}              & 86.9 & 42.0 \\
    \textbf{3s-STJD-CL} (ours)             &  \textbf{91.4}  &  \textbf{61.7} \\

  % \bottomrule
  \end{tabular}

  \label{tab05}
\end{table}

\subsection{Ablation Studies}
\label{ablation}
Ablation studies are performed on the NTU RGB+D 60 dataset using the X-view protocol.

% \subsection{Classification: prime joints vs actionlets} 
% \textbf{Classification: prime joints vs actionlets:} 
% The accuracy of action recognition using actionlets is compared with the accuracy achieved using prime joints detected via STJD, as shown in Fig.~\ref{fig:3}. The accuracy for prime joints is higher than the actionlet, proving that STJD effectively removed more irrelevant information than the actionlet in~\cite{actionlet}. 

% \begin{figure}[tb]
%   \centering
  
% 	\includegraphics[scale=0.4]{fig3.png}
% 	\caption{\textbf{Comparison of action recognition accuracy on the NTU RGB+D 60 dataset using prime-joints, action-let and non-prime Joints.}}
 
%   \label{fig:3}
% \end{figure}

% \subsection{Visualization of prime joints}

\begin{figure}[htb]
    \centering
    \begin{subfigure}{0.4\textwidth}
    \centering
        \includegraphics[width=0.8\linewidth]{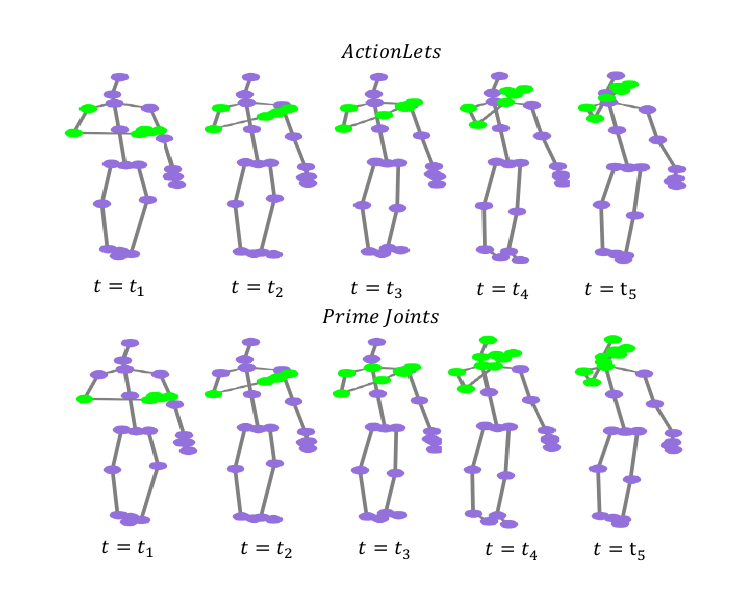}
        \caption{Actionlets vs prime joints for action "Drinking" }
        \label{fig6:sub1}
    \end{subfigure}
    \hfill
    \begin{subfigure}{0.4\textwidth}
    \centering
        \includegraphics[width=0.8\linewidth ]{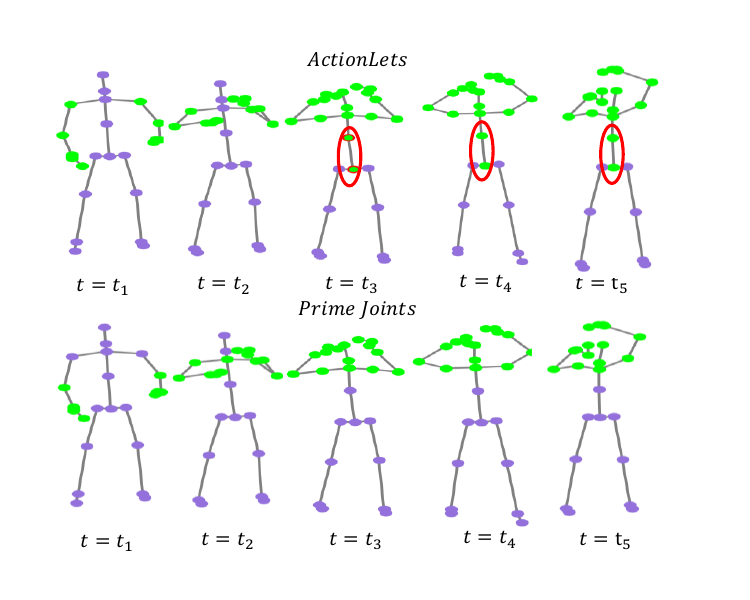}
        \caption{ Actionlets Vs prime joints for action "Comb Hair"  }
        \label{fig6:sub2}
    \end{subfigure}
\caption{ \textbf{Visualization of actionlet and prime joints:} The green joints represent actionlet or prime joints in the respective sequence, and the purple joints belong to non-actionlet or non-prime joints. The {\color{red}\huge 0} highlights the irrelevant joints included in Actionlet.}
    \label{fig6}
\end{figure}

\subsubsection{Visualization of prime joints}
The prime joints detected using STJD are qualitatively compared with Actionlets as shown in Figure~\ref{fig6}.  For action \enquote{Drinking} in \ref{fig6:sub1}, actionlet ignores the discriminative head joints, but they are included in the prime joints. Moreover, for action \enquote{Comb hair} in \ref{fig6:sub2}, irrelevant hip and spine joints are included in the actionlet, but prime joints accurately detected the relevant joints only.

\begin{figure*}
    \centering
    \begin{subfigure}{0.3\textwidth}
        \includegraphics[width=\textwidth]{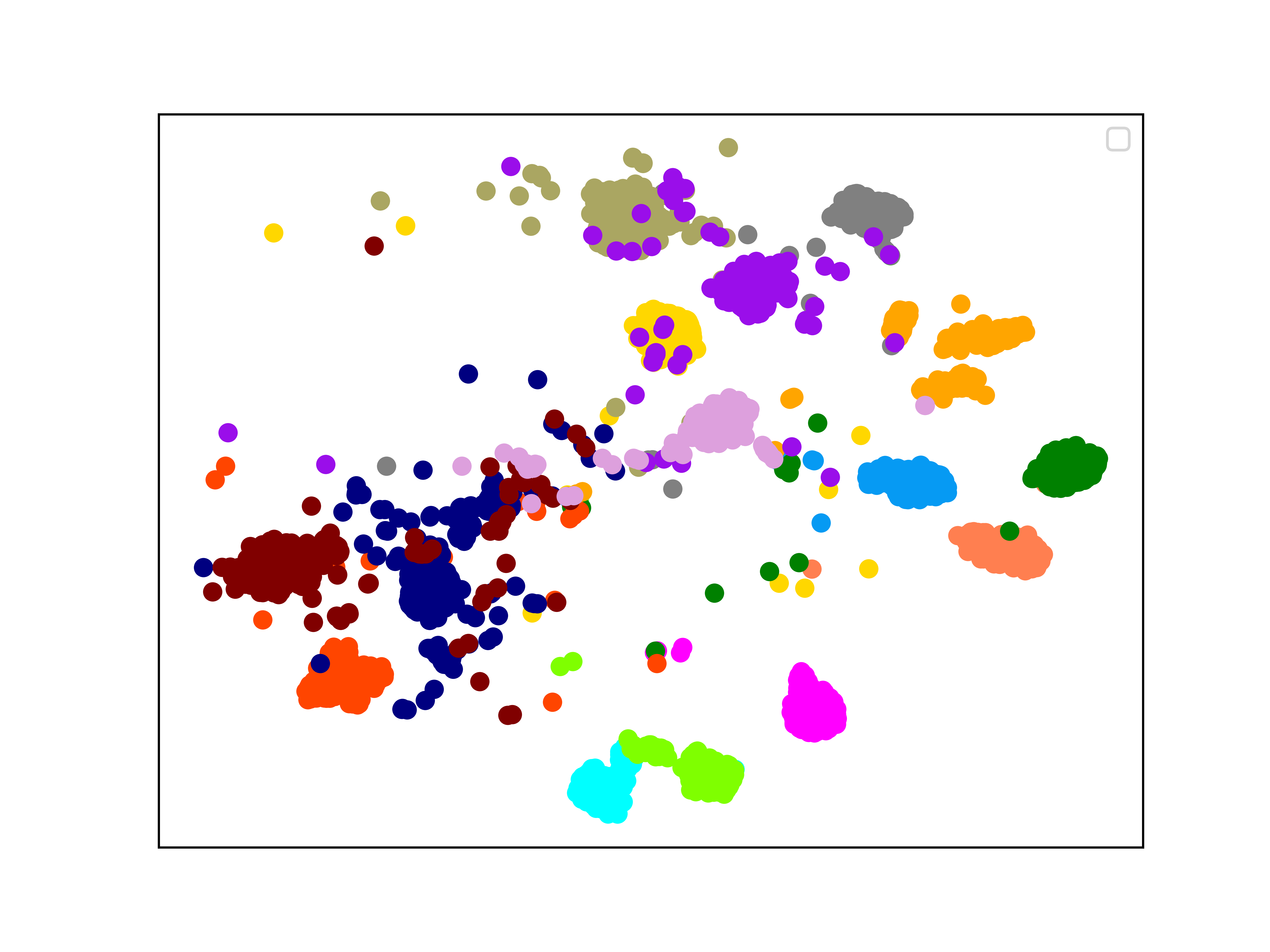}
        \caption{STJD-CL joint  }
        \label{fig5:sub1}
    \end{subfigure}
    \hfil
    \begin{subfigure}{0.3\textwidth}
        \includegraphics[width=\textwidth]{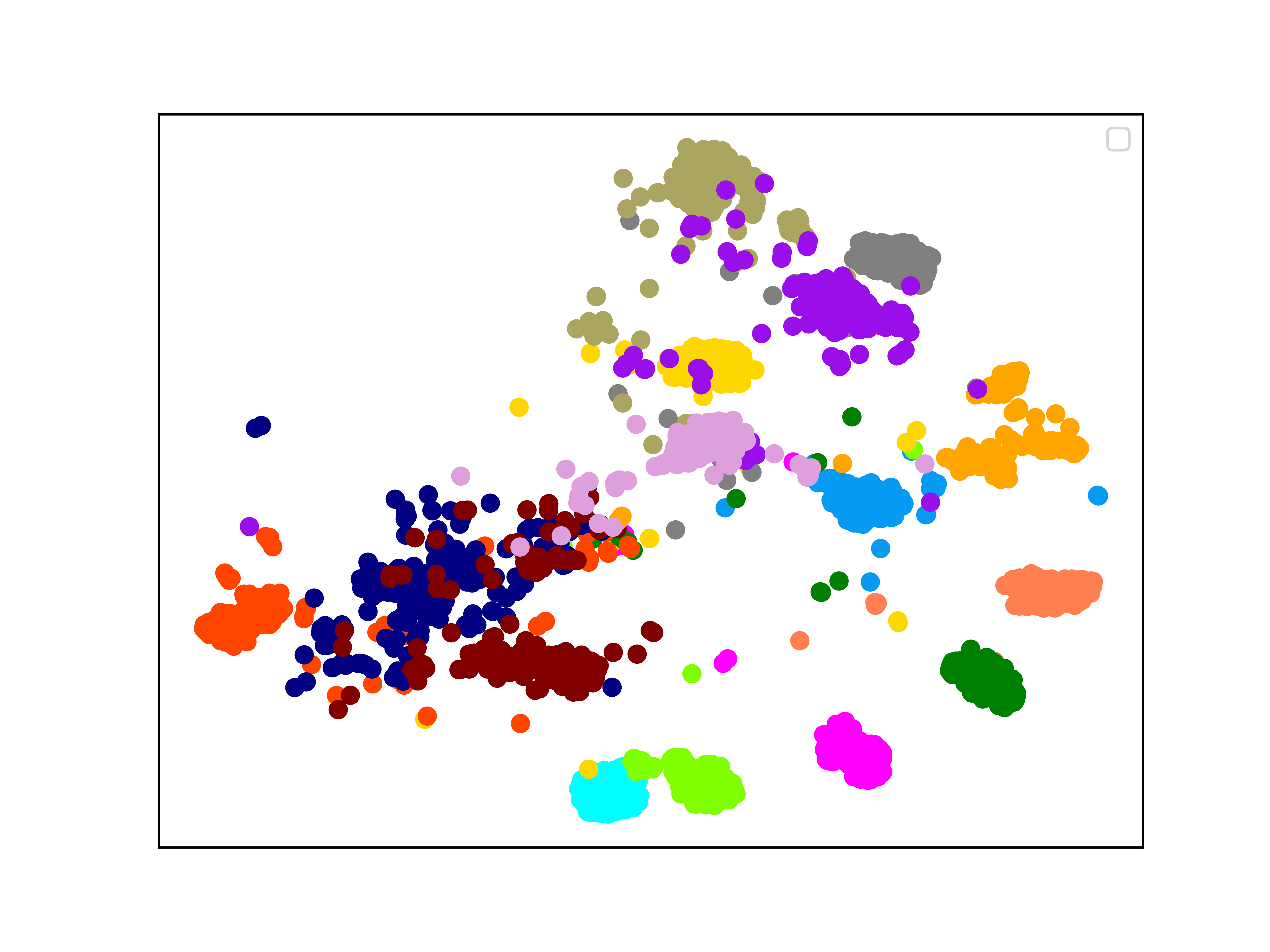}
        \caption{STJD-CL bone }
        \label{fig5:sub2}
    \end{subfigure}
    \hfil
    \begin{subfigure}{0.3\textwidth}
        \includegraphics[width=\textwidth]{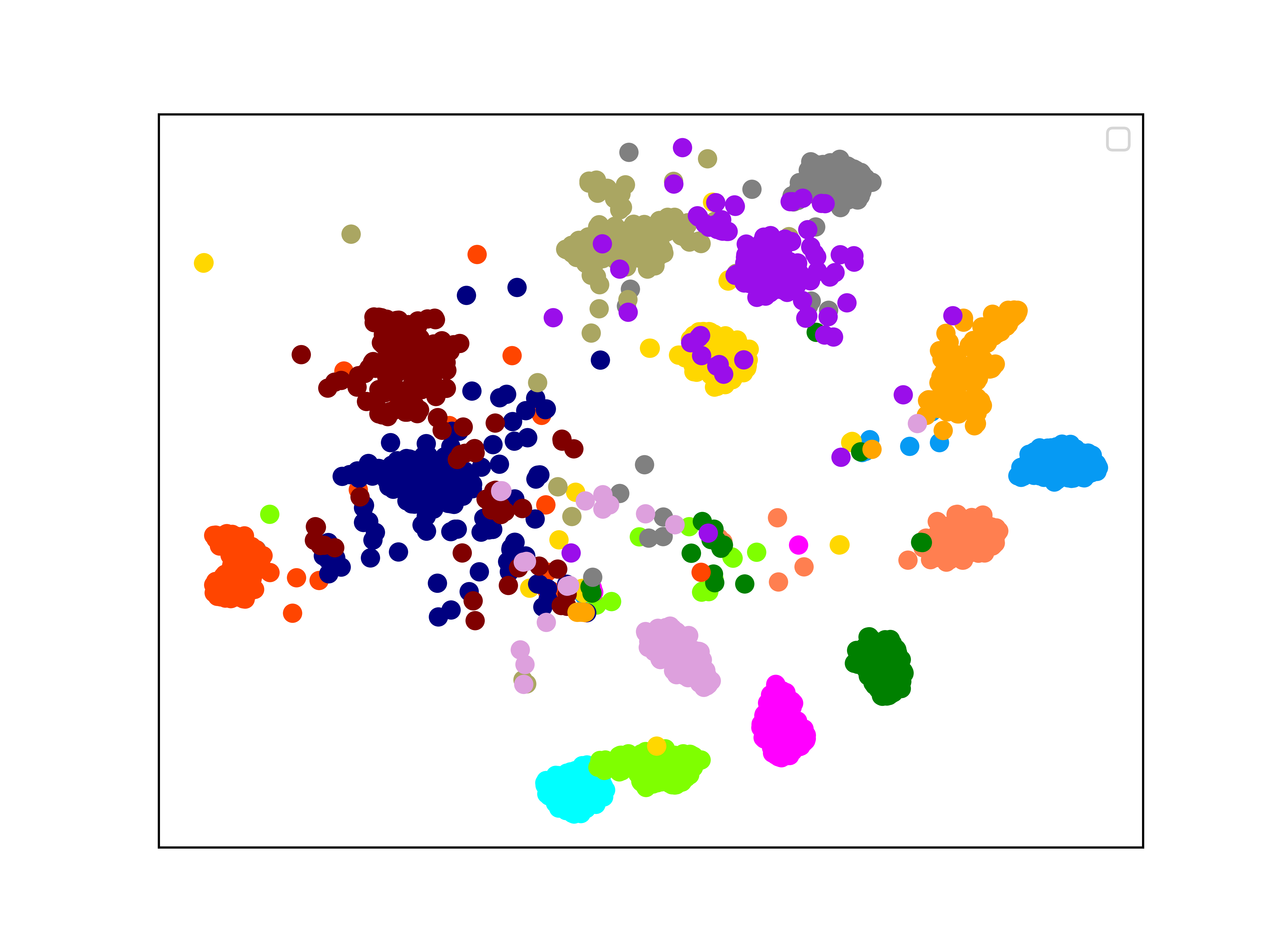}
        \caption{STJD-CL motion}
        \label{fig5:sub3}
    \end{subfigure}

    \medskip
  \begin{subfigure}{0.3\textwidth}
    \includegraphics[width=\textwidth]{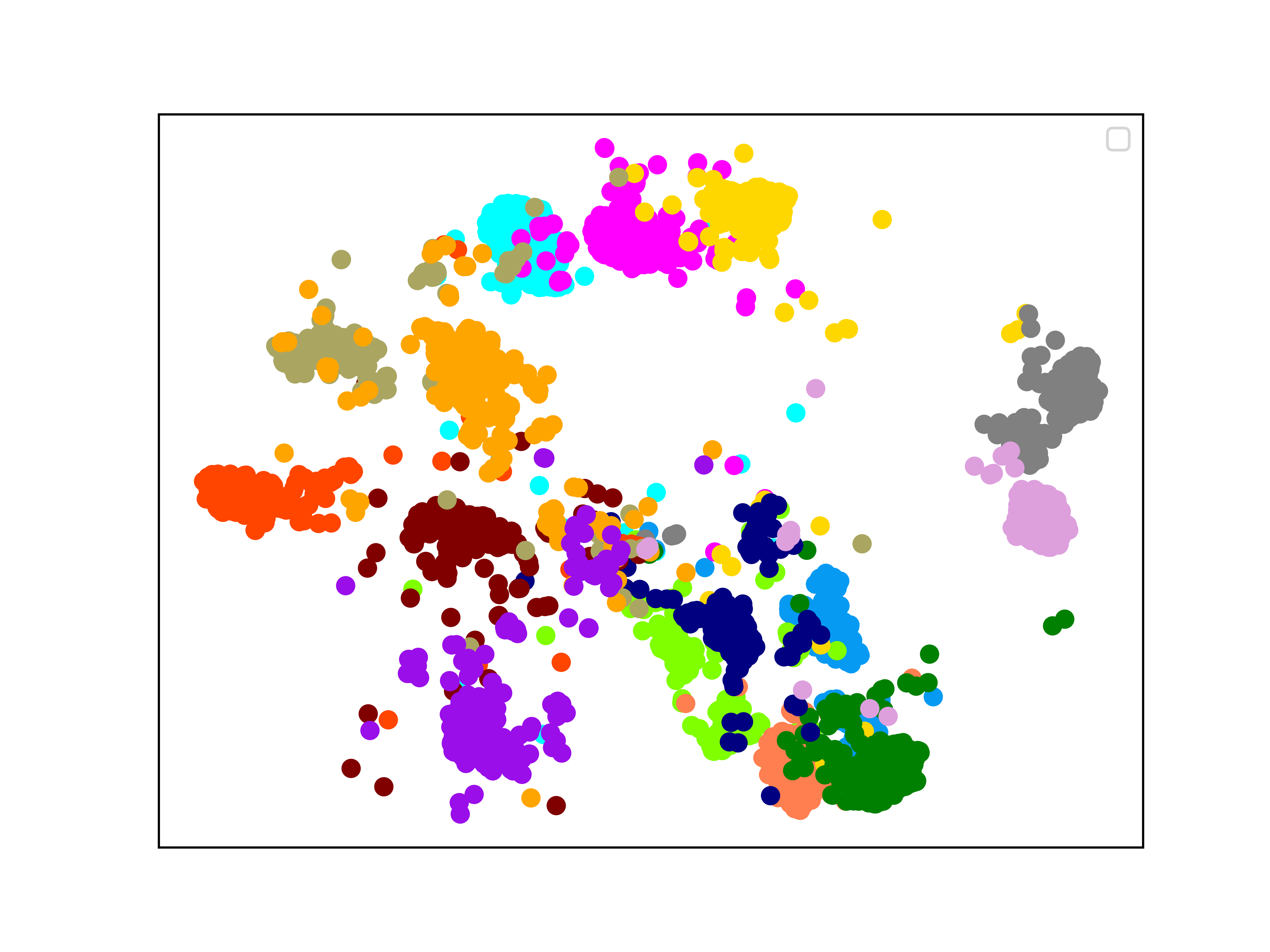}
    \caption{ActCLR Joint}
    \label{fig5:sub4}
  \end{subfigure}
  \hfil
  \begin{subfigure}{0.3\textwidth}
    \includegraphics[width=\textwidth]{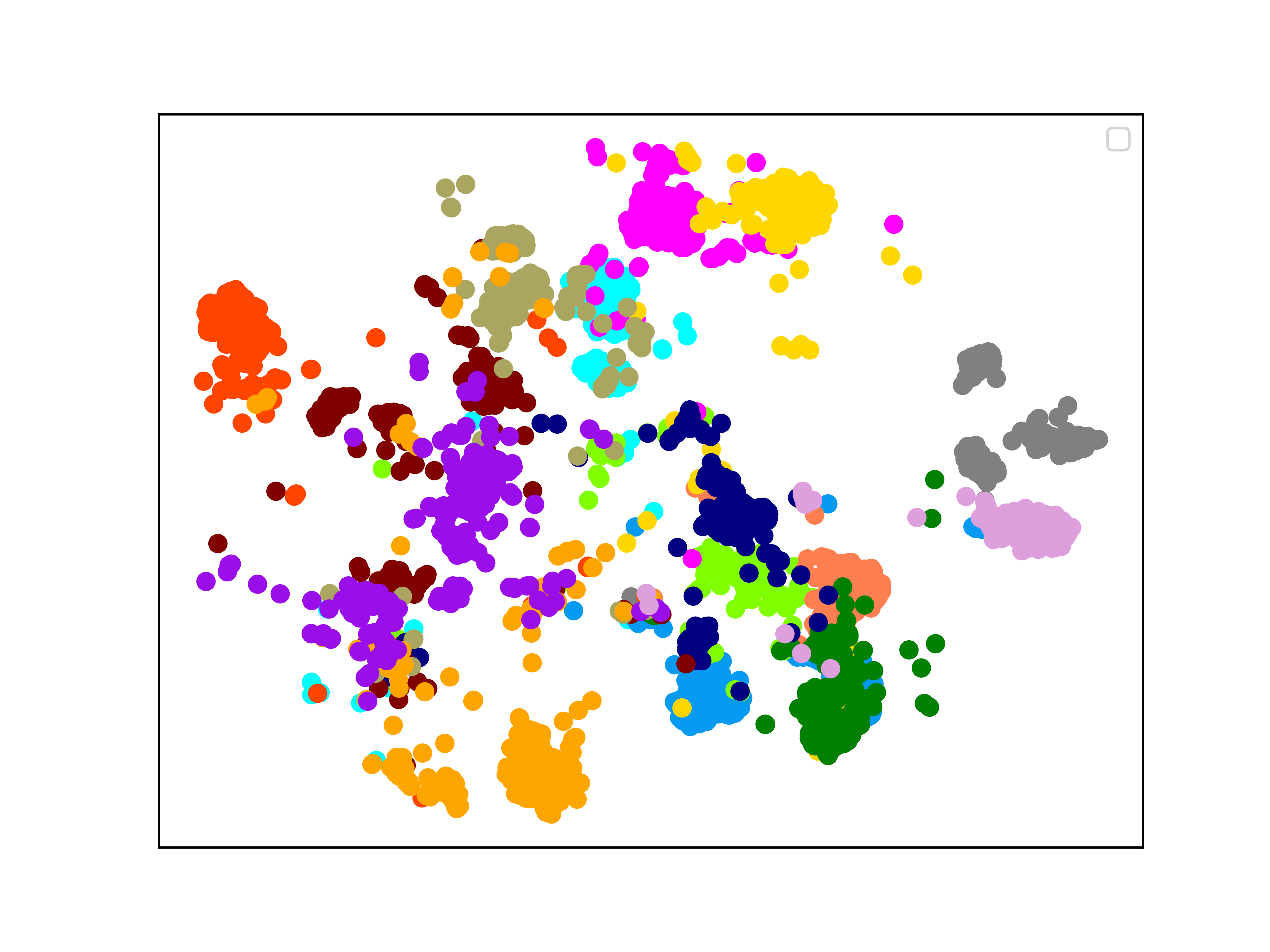}
    \caption{ActCLR bone}
    \label{fig5:sub5}
  \end{subfigure}
  \hfil
  \begin{subfigure}{0.3\textwidth}
    \includegraphics[width=\textwidth]{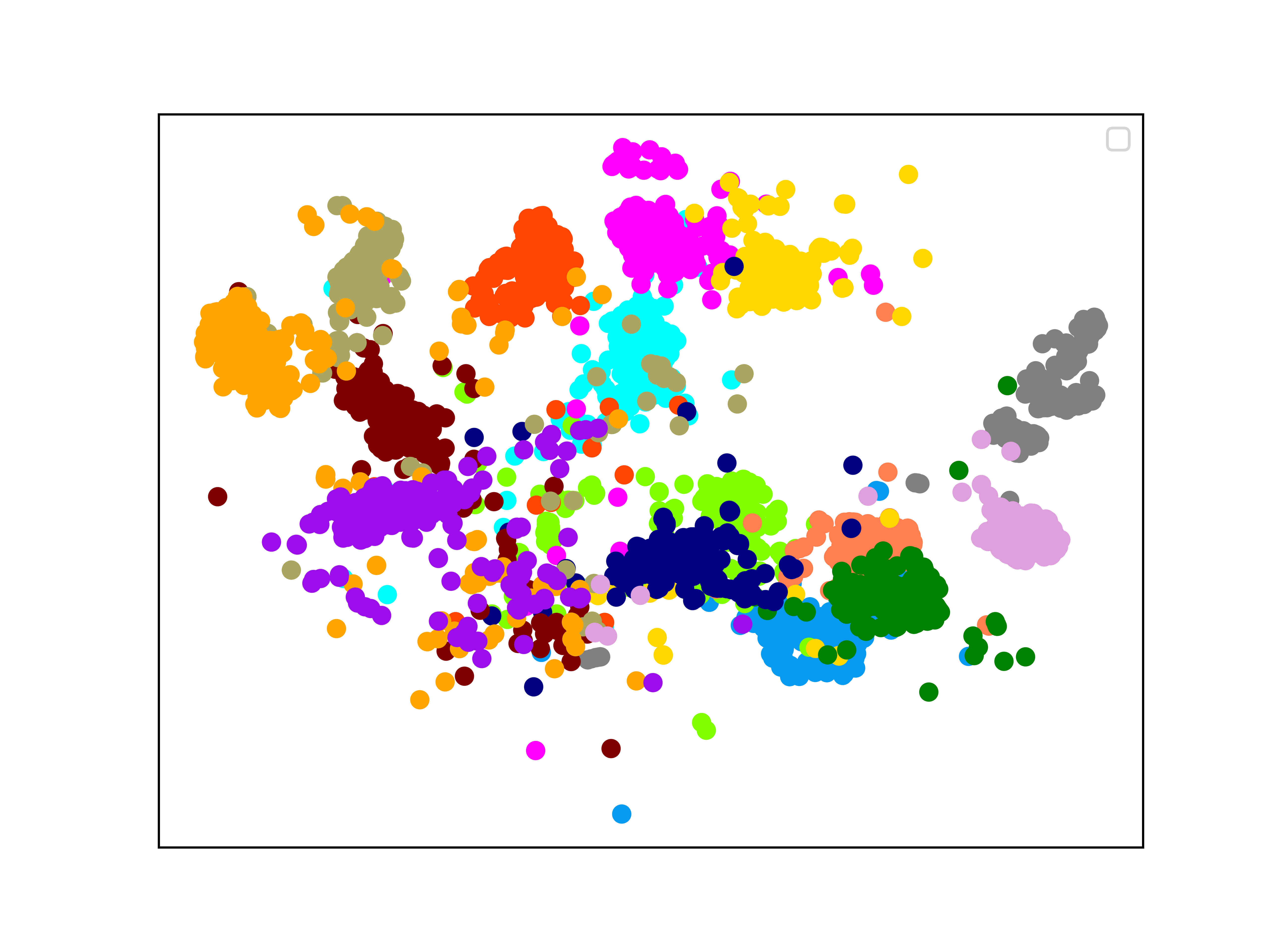}
    \caption{ActCLR motion}
    \label{fig5:sub6}
  \end{subfigure}

\noindent{Legend:} 
{\color{class1}$\blacksquare$} Drinking, 
{\color{class2}$\blacksquare$} Brush Teeth, 
{\color{class3}$\blacksquare$} Decline, 
{\color{class4}$\blacksquare$} Standing, 
{\color{class5}$\blacksquare$} Wearing jacket, 
{\color{class6}$\blacksquare$} Take off jacket, 
{\color{class7}$\blacksquare$} Reach into pocket, 
{\color{class8}$\blacksquare$} Type on the keyboard, 
{\color{class9}$\blacksquare$} Wipe hands, 
{\color{class10}$\blacksquare$} Put the palms together, 
{\color{class11}$\blacksquare$} Contact, 
{\color{class12}$\blacksquare$} Touch chest, 
{\color{class13}$\blacksquare$} Pat each other, 
{\color{class14}$\blacksquare$} Give someone something, 
and {\color{class15}$\blacksquare$} Shake hands.

        \caption{ The t-SNE visualization of embeddings on the NTU RGB+D 60 X-view benchmark. The same randomly selected 15-class samples are used for better clarity. (The ActCLR ~\cite{actionlet} results are obtained by regenerating results with the provided code )  }
    \label{fig5}
\end{figure*}

% \subsection{Embedding space visualization}
\subsubsection{Embedding space visualization}
The embedding spaces obtained using ActCLR~\cite{actionlet} and STJD-CL are visualized in Figure~\ref{fig5}. For clear visualization, the figures showcase the embedding spaces of the same randomly selected 10 action classes from the NTU RGB+D 60  X-View benchmark. As seen, STJD-CL consistently improves the compactness of feature representation within the same class and enhances differentiation across different classes.

\subsubsection{Effectiveness of selecting Prime joints over moving parts}
    To further analyse the advantages of selecting prime joints (action-related moving and static joints) over pre-defined moving body parts, the linear evaluation results were compared using prime joints, prime parts that are covered by any of the prime joints and actionlets. Note that five body parts: left hand, right hand, left leg, right leg, and torso, are pre-defined as in ActCLR~\cite{actionlet}.  Linear evaluation accuracies for the joint stream on the NTU RGB+D 60 dataset are presented in  Table~\ref{tab06}. As seen, the performance of using prime-parts and actionlets are comparable. However, the accuracy improves by about $2$ percentage points on average with prime joints by showing the effectiveness of select action-related moving and static joints.

% \begin{table}[ht]
%     \centering
    
%     \begin{tabular}{l|c|cc|cc}
%         % \toprule
%          & &\multicolumn{2}{c|}{X-view} & \multicolumn{2}{c}{X-sub} \\
         
%         % \cmidrule(lr){3-6}
%         Model& Selection &  KNN & Linear & KNN & Linear \\
%         % \midrule
%        \cline{1-6}
%         ActCLR  & Moving Parts &   78.06 & 86.68  & 73.74 &80.91 \\
%         STJD-CL & Moving and static Parts                & 78.15 &86.82  & 73.59 & 80.83\\
%         STJD-CL  &Moving and static Joints           &\textbf{80.35} & \textbf{87.93} & \textbf{76.08} & \textbf{82.31}\\
%         % \midrule

%         % \bottomrule
%     \end{tabular}
%     \caption{The comparison of classification using prime joints, prime parts and actionlets}
%     \label{tab06}
% \end{table}

\begin{table}[htb]
    \caption{The comparison of linear classification using prime joints, prime parts and actionlets}
    \centering
    
    \begin{tabular}{l|c|cc}
        % \toprule
         % & & X-sub &X-view  \\
         
        % \cmidrule(lr){3-6}
        Model& Selection & X-sub &X-view  \\
        % \midrule
       \cline{1-4}
        ActCLR  & Moving Parts &   80.91 & 86.68   \\
        STJD-CL & Moving and static Parts                & 80.83 &86.82  \\
        STJD-CL  &Moving and static Joints  &\textbf{82.31} & \textbf{87.93} \\
        % \midrule

        % \bottomrule
    \end{tabular}

    \label{tab06}
\end{table}

\subsubsection{Ablation on $L_{RCL}$ loss} The effect of $L_{RCL}$ is validated by removing it from the objective function. Linear evaluation accuracies on NTU RGB+D using only joint modality are presented in Table \ref{tab07}. The incorporation of $L_{RCL}$ yielded a 0.36 percentage point improvement on X-view compared to the STJD-CL without $L_{RCL}$, highlighting its contribution in guiding the learning process.

\begin{table}
    \caption{The comparison of the performance with and without $L_{RCL}$ in the objective function.}
    \centering
    
    \begin{tabular}{l|c|c|c c|}
        % \toprule
         & & & \multicolumn{2}{c|}{Linear Eval}  \\
         
        % \cmidrule(lr){3-6}
        Model & $L_{CL}$  &$L_{RCL}$ &   X-view &  X-sub \\
        % \midrule
       \cline{1-5}
        % ActCLR & $\checkmark$  & & 86.7 & 80.9\\
         STJD-CL& $\checkmark$  &       &    87.63     &    81.95   \\
         STJD-CL& $\checkmark$ &  $\checkmark$  &    87.93     &     82.31  \\
       
   % \midrule

        % \bottomrule
    \end{tabular}

    \label{tab07}
\end{table}

\subsubsection{Statistical Validation of STJD Performances}

A paired t-test was conducted to statistically validate the significance of the performance improvement by the proposed STJD. A linear classifier was trained on frozen encoder and evaluated five times, each time with a randomly initialized classifier, following standard evaluation practices~\cite{t_test2}.

The performance of STJD-CL was compared to that of ActCLR using the t-test. The null hypothesis (\(H_0\)) assumes no significant difference in their performances $H_0: \mu_{STJD} = \mu_{ActCLR}$ , while the alternative hypothesis (\(H_1\)) states that there is a significant difference $H_1: \mu_{STJD} \neq \mu_{ActCLR}$. Here, $\mu_{STJD}$ and  $\mu_{ActCLR}$ represent the true mean performance of the respective models. With 4 degrees of freedom and a significance level of \( \alpha = 0.05 \), the null hypothesis was rejected if the computed t-value exceeded the critical value.

The results, which are presented in Table~\ref{t_test_results}, indicate that a statistically significant difference was observed between STJD-CL and ActCLR. It was further confirmed that the true mean performance of STJD-CL was higher than that of ActCLR in both the X-view and X-sub evaluation protocols, thereby demonstrating that STJD-CL outperformed the ActCLR method and the performance difference is statistically significant.

\begin{table}[htbp]
\centering
\scriptsize
\caption{Paired t-test results on NTU RGB+D X-view and X-sub evaluations.}
\begin{tabularx}{\columnwidth}{l|X|X|X|X|X}
    \hline
    Test & $\mu_{\text{STJD-CL}}$ & $\mu_{\text{ActCLR}}$ & Computed t-value & Critical t-value at $\alpha=0.05$ & Decision \\
    \hline
    X-view & 87.92 & 86.71 & 79.23 & $\mp 5.598$ & Reject $H_0$ \\
    X-sub  & 82.30 & 80.90 & 95.30 & $\mp 5.598$ & Reject $H_0$ \\
    \hline
\end{tabularx}
\label{t_test_results}
\end{table}

% \begin{table}[h]
%     \centering
%     \caption{Paired t-test results on NTU RGB+D X-view and X-sub evaluations.}
%     \begin{tabular}{l|c|c|c|c}
%         \hline
%          &  Computed t-value & \multicolumn{1}{c|}{Critical t-value at} \\ 
%              &                 & \multicolumn{1}{c|}{\( \alpha = 0.05 \)} & Decision \\
%         \hline
%         X-view  &79.23  & $\pm$ 5.598 & Reject \( H_0 \) \\
%         X-sub  & 95.3 & $\pm$ 5.598 & Reject \( H_0 \) \\
        
%         \end{tabular}
%     \label{t_test_results}
% \end{table}

\section{Discussion and conclusion}
\label{discussion}
This paper introduces the Spatio-Temporal Joint Density (STJD), a novel measurement for quantifying the joint motion and intricate interactions between moving and static joints simultaneously for skeleton-based action recognition. Using STJD, a set of discriminative prime joints is detected, and these prime joints effectively enhance learning in both contrastive and reconstructive frameworks, namely STJD-CL and STJD-MP, as demonstrated by the experiments. 
%The proposed STJD-CL is then employed to leverage these prime joints for guiding contrastive learning, while STJD-MP is used to mask prime joints, thereby challenging the decoder to learn more discriminative representations at the encoder.Compared to existing state-of-the-art methods, both STJD-CL and STJD-MP demonstrate superior performance in learning effective representations for various downstream tasks.

While STJD-CL and STJD-MP improved recognition accuracy for the most challenging actions compared to ActCLR~\cite{actionlet} and MAMP~\cite{mao01}, they experienced a slight accuracy drop in linear and fine-tune evaluations for a few actions, such as {Playing with a mobile phone} and  \textquote{Type on the key board}. These actions are performed with the hands, and once the hands move to a particular position, subtle movements of the hand joints do not lead to significant changes in STJD. Moreover, in certain poses, the STJD calculations become confused, leading to misdetections. For instance, in the action "Take off shoes," posture variations may induce density changes in irrelevant joints, such as the spine or hip, which are then incorrectly identified as prime joints. This limitation affects the detection of the prime joints for these action instances. This issue is expected to be mitigated by extending the concept of prime joints to prime parts and using them adaptively. This will be our future improvement to the proposed method.

\bibliographystyle{unsrt}

\bibliography{bibliography}


\begin{IEEEbiography}[{\includegraphics[width=1in,height=1.25in,clip,keepaspectratio]{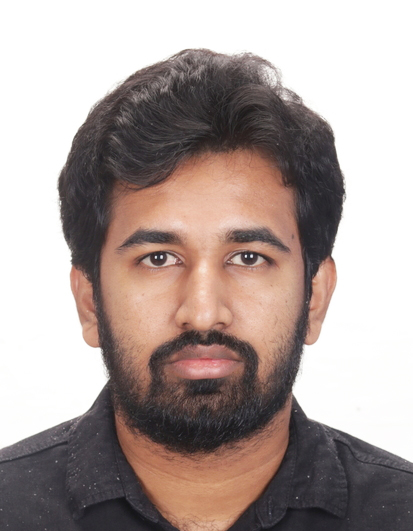}}]{Shanaka Ramesh Gunasekara}  (Member, IEEE) received the B.Sc. (hons) degree in electrical and electronic engineering from the University of Peradeniya, Sri Lanka. He is currently pursuing the Ph.D. degree with the Advanced Multimedia Research Lab (AMRL), University of Wollongong, Australia, with a focus on human action recognition. His research interests include 3D computer vision, human motion analysis, signal processing, medical image analysis, and robotics.\end{IEEEbiography}

\begin{IEEEbiography}[{\includegraphics[width=1in,height=1.25in,clip,keepaspectratio]{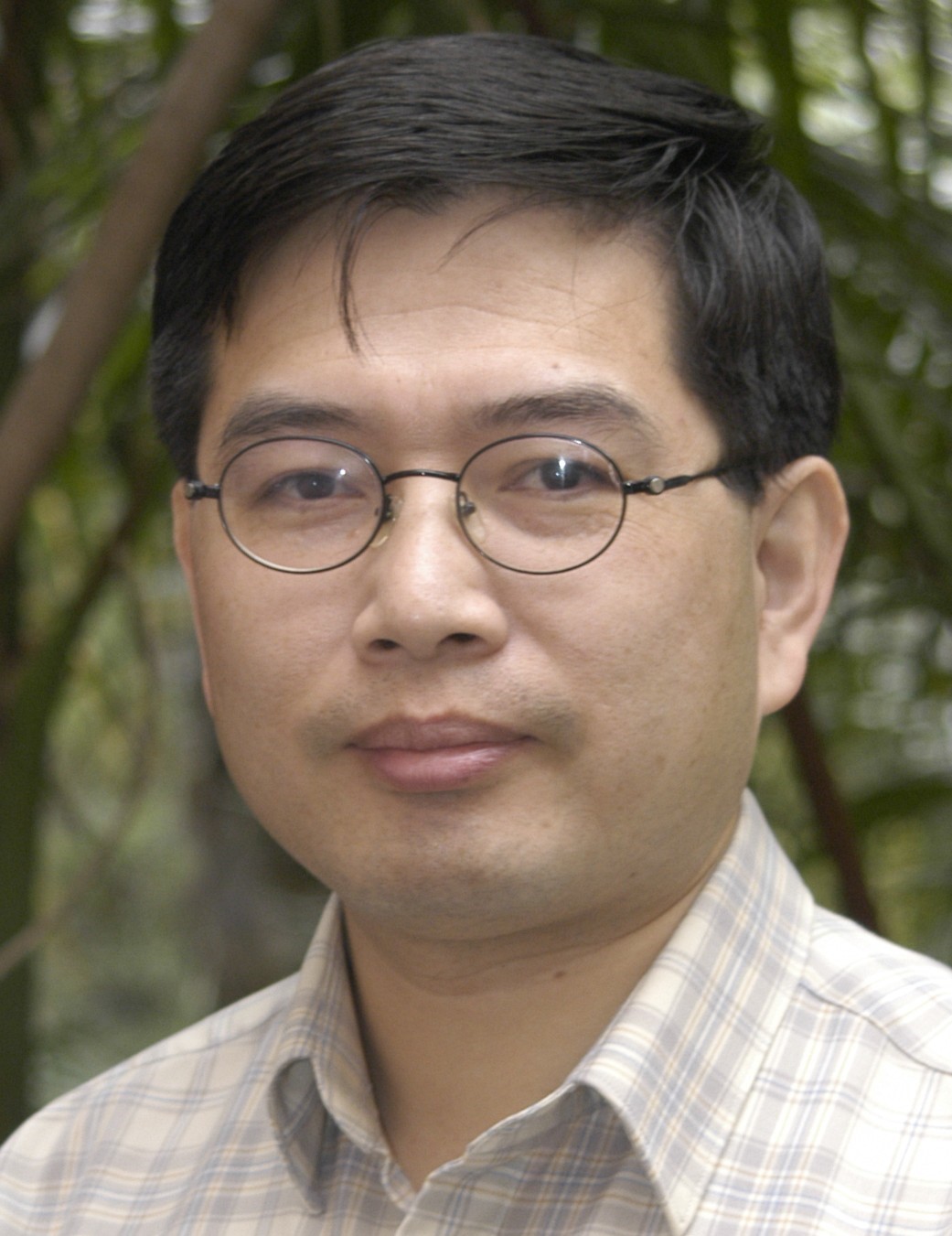}}]{Wanqing Li} (M’97-SM'05) received his PhD in electronic engineering from the University of Western Australia. He was a Senior Researcher and later a Principal Researcher at the Motorola Research Lab in Sydney from 1998 to 2003, and a visiting researcher at Microsoft Research, USA, in 2008, 2010, and 2013. He is currently a Professor and Co-Director of the Advanced Multimedia Research Lab (AMRL), University of Wollongong, Australia. His research areas include machine learning, 3D computer vision, 3D multimedia signal processing, medical image analysis, natural language processing, and their applications.
Dr. Li served as a Technical Program Co-Chair for IEEE ICME 2021 and has served as Co-Chair for many IEEE Workshops. He is an Associate Editor for IEEE Transactions on Image Processing and IEEE Transactions on Multimedia. He served as an Associate Editor for IEEE Transactions on Circuits and Systems for Video Technology from 2018 to 2021 and for the Journal of Visual Communication and Image Representation from 2016 to 2019.\end{IEEEbiography}

\begin{IEEEbiography}[{\includegraphics[width=1in,height=1.25in,clip,keepaspectratio]{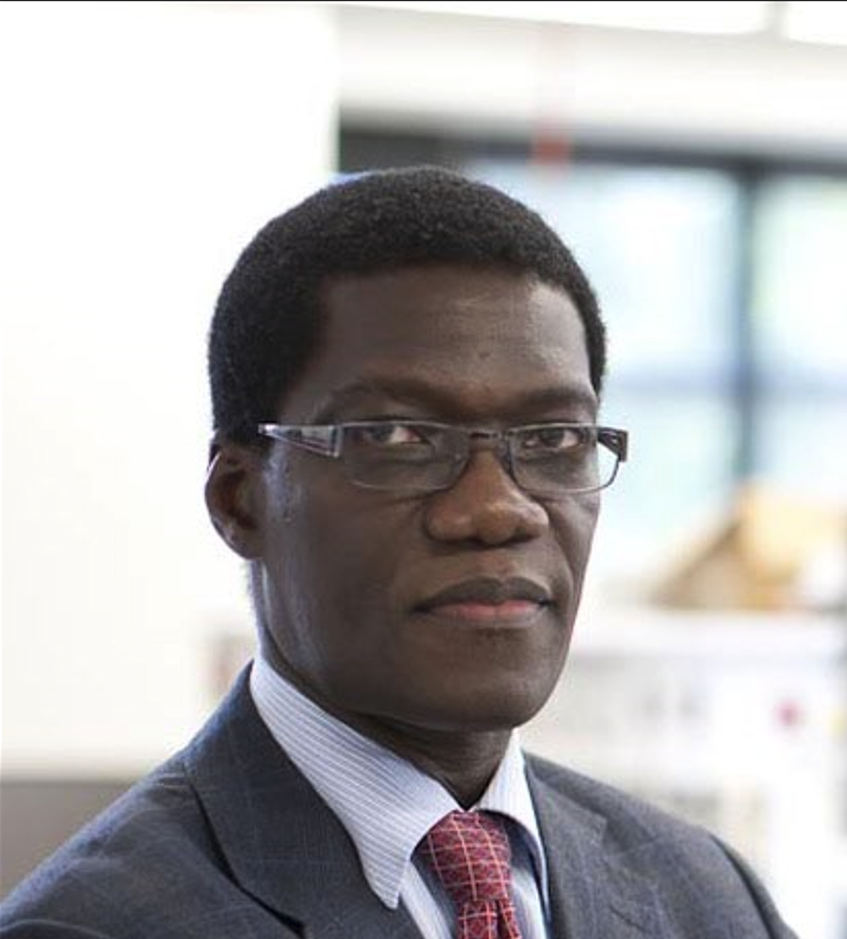}}]{Philip O. Ogunbona }  received the B.Sc. degree (with first class honours) in electronics and electrical engineering from the University of Ife, Nigeria, and the Ph.D. degree in electrical engineering from Imperial College London, U.K. He is a professor in computer science at the University of Wollongong, Australia. His research interests include signal and image processing, machine learning, computer vision and natural language processing. Professor Ogunbona is Fellow of the Australian Computer Society and a Life Senior Member of IEEE.\end{IEEEbiography}

\begin{IEEEbiography}[{\includegraphics[width=1in,height=1.25in,clip,keepaspectratio]{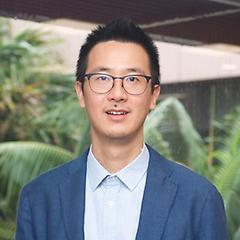}}]{Jie (Jack) Yang } is a lecturer of Big Data Analytics in the School of Computing and Information Technology at the University of Wollongong. His major research interests include Data Mining and Natural Language Processing. Dr. Yang has been the Chief Investigator for three research grants of Discovery/Linkage Project themes from the prestigious Australian Research Council (ARC) and has published over 70 articles.\end{IEEEbiography}

\newpage


%\vfill


\end{document}